\pdfoutput=1

\documentclass[11pt]{article}

\usepackage[preprint]{acl}

\usepackage{times}
\usepackage{latexsym}
\usepackage{multicol}
\usepackage{multirow}
\usepackage{makecell}
\usepackage{booktabs}
\usepackage{graphicx}
\usepackage{algorithm}
\usepackage{algorithmic}
\usepackage{amsmath}
\usepackage{enumitem}
\usepackage{longtable}
\usepackage{subcaption}
\usepackage{colortbl}
\usepackage{xcolor}
\usepackage{soul}

\usepackage{tcolorbox}
\tcbuselibrary{raster}
\tcbuselibrary{breakable}
\usepackage{listings}
\usepackage{xcolor}

\definecolor{codegreen}{rgb}{0,0.6,0}
\definecolor{codegray}{rgb}{0.5,0.5,0.5}
\definecolor{codepurple}{rgb}{0.58,0,0.82}
\definecolor{backcolour}{rgb}{0.95,0.95,0.92}

\lstdefinestyle{mystyle}{
    backgroundcolor=\color{backcolour},   
    commentstyle=\color{codegreen},
    keywordstyle=\color{magenta},
    numberstyle=\tiny\color{codegray},
    stringstyle=\color{codepurple},
    basicstyle=\ttfamily\small,
    breakatwhitespace=false,         
    breaklines=true,                 
    captionpos=b,                    
    keepspaces=true,                 
    numbersep=5pt,                  
    showspaces=false,                
    showstringspaces=false,
    showtabs=false,                  
    tabsize=2
}

\usepackage{tikz}
\usetikzlibrary{shapes}
\newcommand{\inlinecode}[1]{%
    \begin{tikzpicture}[baseline=0ex]%
         \node[anchor=base,%
         text height=0.7em,%
         text depth=0.7ex,%
         inner ysep=0pt,%
         draw=lightgray!50,%
         fill=lightgray!50,%
         rounded corners=2pt] at (0,0) {\footnotesize\texttt{#1}};%
    \end{tikzpicture}%
}

\newcommand{\ctext}[3][RGB]{%
  \begingroup
  \definecolor{hlcolor}{#1}{#2}\sethlcolor{hlcolor}%
  \hl{#3}%
  \endgroup
}

\newcommand{\inlinecodesmall}[1]{%
    \begin{tikzpicture}[baseline=0ex]%
         \node[anchor=base,%
         text height=0.7em,%
         text depth=0.7ex,%
         inner ysep=0pt,%
         draw=lightgray!50,%
         fill=lightgray!50,%
         rounded corners=2pt] at (0,0) {\tiny \texttt{#1}};%
    \end{tikzpicture}%
}

\usepackage[T1]{fontenc}

\usepackage{xspace}
\usepackage[utf8]{inputenc}
\newcommand{\ourobjective}[0]{\texttt{AssistV}\xspace}

\usepackage{microtype}

\usepackage{inconsolata}

\title{Learning Task Decomposition to Assist Humans\\in Competitive Programming}

\author{
  Jiaxin Wen$^{1,2}$,
  Ruiqi Zhong$^3$,
  Pei Ke$^{1,2,\dagger}$,
  Zhihong Shao$^{1,2}$,\\
  \textbf{Hongning Wang}$^{1,2}$
  \textbf{, Minlie Huang}$^{1,2,\dagger}$ \\
 $^1$The CoAI group, Tsinghua University, Beijing, China \\
  $^2$Department of Computer Science and Technology, Tsinghua University, Beijing, China\\
  $^3$ University of California, Berkeley \\
  \texttt{wenjx22@mails.tsinghua.edu.cn, aihuang@tsinghua.edu.cn} \\
}

\begin{document}
\maketitle

\begin{abstract}
When using language models (LMs) to solve complex problems, humans might struggle to understand the LM-generated solutions and repair the flawed ones. 
To better assist humans in repairing them, we propose to automatically decompose complex solutions into simpler pieces corresponding to specific subtasks. We introduce a novel objective for learning task decomposition, termed \emph{assistive value} (\ourobjective), which measures the feasibility and speed for humans to repair the decomposed solution. We collect a dataset of human repair experiences on different decomposed solutions. Utilizing the collected data as in-context examples, we then learn to critique, refine, and rank decomposed solutions to improve \ourobjective. We validate our method under competitive programming problems: under 177 hours of human study, our method enables non-experts to solve 33.3\% more problems, speeds them up by 3.3x, and empowers them to match unassisted experts.
\end{abstract}

\section{Introduction}
{\let\thefootnote\relax\footnotetext{$^\dagger$ Corresponding author}}

With their increased capabilities, language models (LMs) are used to perform increasingly complex and high-impact problems \citep{trinh2024solving, li2022competition, huang2023benchmarking}. 
This however causes the scalable oversight challenge \citep{amodei2016concrete}:
LMs might fail to provide reliable solutions for these problems, but it is also difficult for humans to evaluate and improve LMs' solutions due to the required significant time and expertise.

\begin{figure*}
    \centering
    \includegraphics[width=\linewidth]{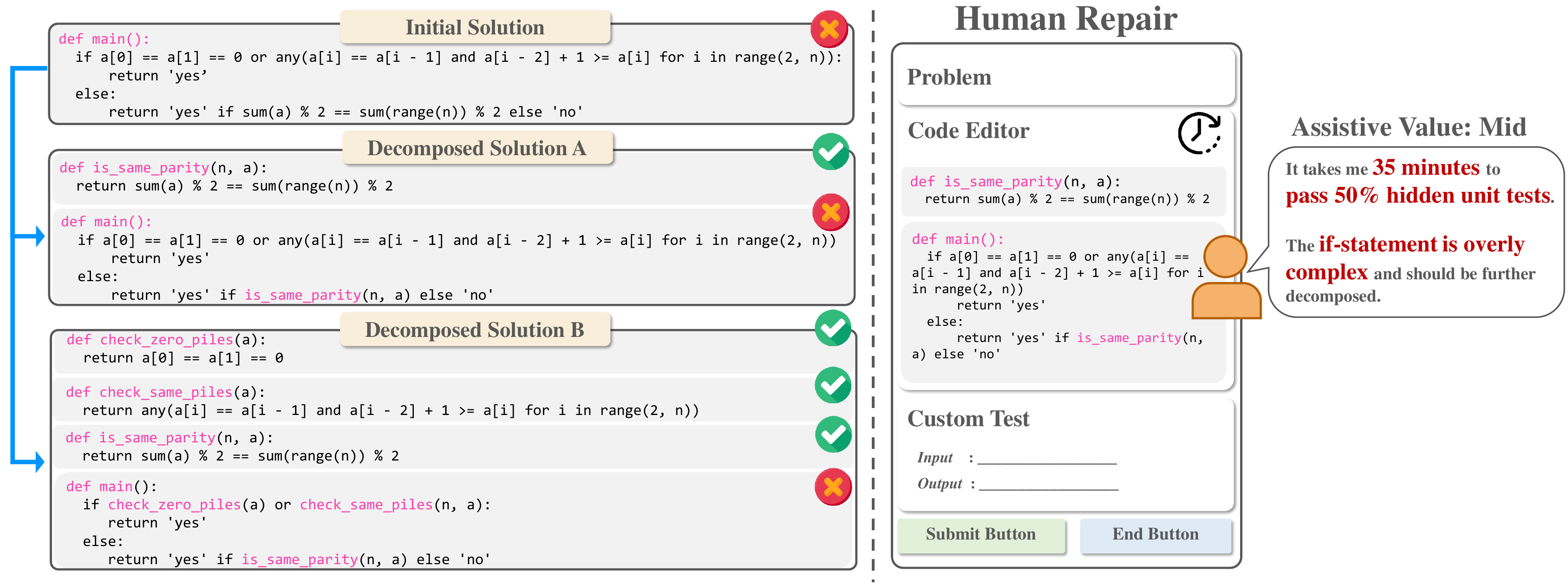}
    \caption{Decompositions can assist humans in supervising models to solve complex problems. \textbf{Left}: To solve a problem, an LM would first propose an initial solution; our goal is to decompose the initial solution into multiple simpler pieces such that humans can repair it more easily. (Sub)Task descriptions are truncated for brevity. \textbf{Right}: The \emph{assistive value} (\ourobjective) of a decomposition measures the feasibility and speed of humans to repair the decomposed solution in the actual problem-solving process. For example, Decomposition B has a higher \ourobjective value than A in practice, as it further decomposes the complex if-statement into two simpler subtasks, which effectively assists humans in identifying a missing condition.}
    \label{fig:intro}
\end{figure*}

One strategy to assist humans is task decomposition: as shown in Figure \ref{fig:intro}, humans can more easily understand and repair complex solutions after they are decomposed into simpler pieces that correspond to specific subtasks \citep{lee2001does}. 

However, not all decompositions are helpful, and it is challenging for humans to design a generally effective one \cite{connolly1997decomposed,selby2015relationships, correa2023humans}.
For example, \citet{charitsis2023detecting} showed that improper decomposition designed by novice programmers can impede human debugging performance. 
To decompose better, we need methods beyond using fixed heuristic rules \cite{wu2021recursively} or learning from author-crafted demonstrations \cite{yao2022react,zelikman2023parsel}.

In this paper, we introduce a novel objective for learning task decomposition: \emph{assistive value} (\ourobjective, Eq \eqref{equation:efficiency}), 
which measures the feasibility and speed of humans to repair a decomposed solution in the actual annotation process (Figure \ref{fig:intro} right). 
To improve \ourobjective, we first collect a dataset of decompositions, measure their \ourobjective, and ask human annotators to provide a natural language critique on what makes a decomposition (not) helpful. 
Then we design a three-stage process to generate high-\ourobjective decompositions, where each stage is implemented by an LLM that learns from our dataset in context: 1) learn a critique model $\pi_\text{critique}$ for predicting human critique on how to improve the initial decomposition for higher \ourobjective, 2) learn a refine model $\pi_\text{refine}$ to incorporate the critique to refine the initial decomposition, and 3) learn a rank model $\pi_\text{rank}$ to select a decomposition with high \ourobjective.

We chose competitive programming as a testbed to validate our method for scalable oversight, since it is a challenging task that both LMs and humans alone struggle to solve. 
We recruit 30 Python programmers, including 11 experts \footnote{This group includes medalists in National or International
Olympiad in Informatics.} and 19 non-experts, to repair model-generated program solutions for competitive coding problems, resulting in a total of 177 worker hours. 
Experiment results show that our method enables humans to solve 33.3\% more problems, speeds up non-experts and experts by 3.3 and 2.4 times, and assists non-experts to match the performance of non-assisted experts. 
We then analyze LMs' ability to select decompositions with higher \ourobjective:
while humans' intuitive judgment is not better than random (49.5\%), GPT-3.5-Turbo achieves 62.5\% accuracy by learning from human repair experiences, and GPT-4 is 15.6\% better. This result indicates that LMs could learn to perform better than humans at predicting what is more helpful for humans.

Our core contributions are:
\begin{itemize}[topsep=0pt, itemsep=-2pt]
    \item We assist humans with scalable oversight via automated task decomposition.
    \item We introduce a novel objective for learning task decomposition: \ourobjective, and we propose a three-stage method to produce a high-\ourobjective decomposition by learning to critique, refine, and rank decompositions.
    \item We show that our method is effective in competitive programming, improving human supervision performance and bridging the expertise gap.
\end{itemize}

Overall, even when LMs cannot solve the problem themselves, they can still learn to assist humans.
By learning from human experiences to repair solutions, more capable models can predict assistive values more accurately and sometimes more accurately than humans, highlighting the potential of learning-based methods to assist humans.

\section{Methodology} \label{section:method}

\begin{figure*}[t!]
    \centering
    \includegraphics[width=\linewidth]{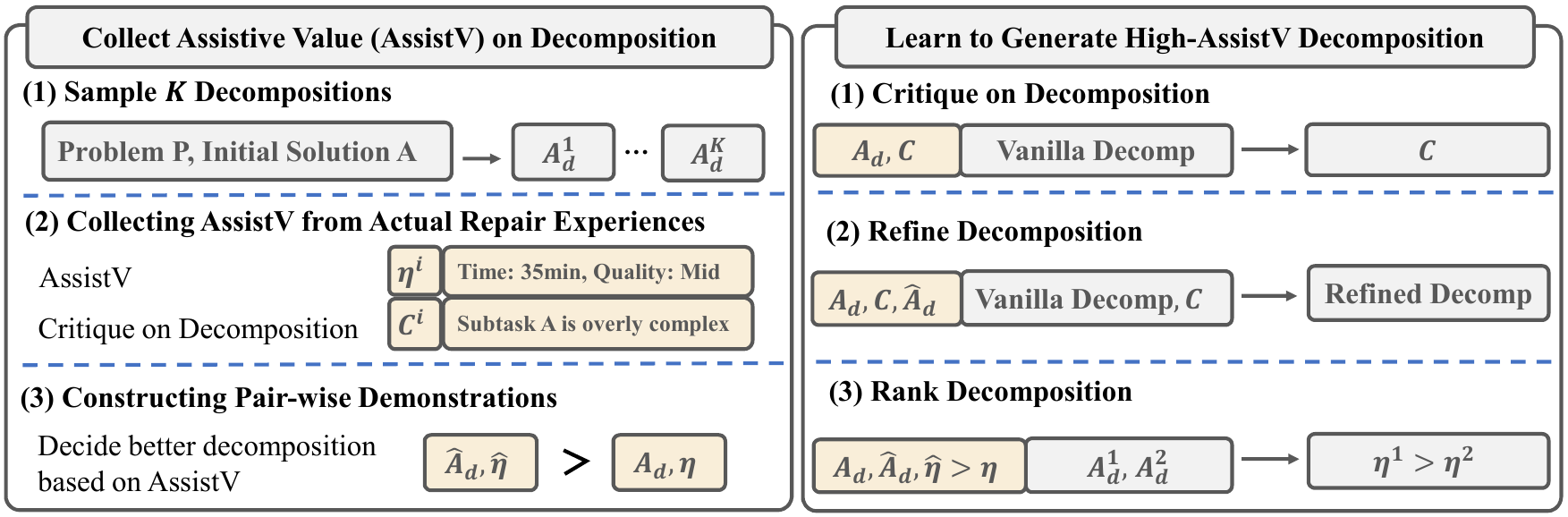}
    \caption{Method overview. \textbf{Left}: we sample multiple decompositions from LMs and evaluate them based on assistive value $\eta$ and critique $C$. We then construct pair-wise decompositions to demonstrate the difference between low- and high-\ourobjective decompositions. \textbf{Right}: Starting from a vanilla decomposition generated by naively prompting LMs, we use the collected pair-wise data as in-context demonstrations to learn three models to critique, refine, and rank decompositions to better assist humans.}
    \label{fig:overview}
\end{figure*}

\subsection{Task Definition} \label{sec:task-definition}
Our objective is to assist human labelers in repairing model-generated solutions to complex tasks by learning a decomposition model. This model decomposes input solutions into multiple easier-to-repair pieces corresponding to specific subtasks.

Formally, given a problem $P$ and an initial model solution $A$, we aim to transform $A$ into a decomposed solution $A_d$ to improve its assistive value $\eta$, as defined by:
\begin{equation}\label{equation:efficiency}
\begin{aligned}
\eta(A_d)= \int_0^T \texttt{eval}(A_d^t)\mathrm{d}t  
\end{aligned}
\end{equation}
where $A_d^t$ denotes the solution repaired by humans after spending time $t$ from its initial solution $A_d$, and $\texttt{eval}(\cdot)$ is a metric of solution quality. In competitive programming, we set $\texttt{eval}(\cdot)$ as the passing rates on unit test cases. Assistive value $\eta(A_d)$ summarizes the solution quality over the human repairing process starting from $A_d$ \cite{bradley1997use}, with example trajectories shown in Figure \ref{fig:human_debugging}. A higher value of $\eta(A_d)$ indicates that assisted humans can more easily repair $A_d$, thus efficiently providing a high-quality label to the problem $P$.

While there are diverse ways to decompose one single solution as shown in Figure \ref{fig:intro}, effective decomposition that improves human problem-solving performance is non-trivial and can even be challenging for humans to devise \cite{charitsis2023detecting}. To decompose better, we need methods beyond prior methods that rely on fixed heuristic rules or learn from author-crafted demonstrations.  In this paper, we propose to improve the assistive value of task decomposition by learning from human problem-solving experiences.

Specifically, we first construct a training set of human repair experiences between different decomposed solutions $D_{train}=\{(\hat{A}_d, A_d, C, P)\}$, where $\eta(\hat{A}_d)>\eta(A_d)$ and $C$ is a natural language critique that explains why the decomposition $A_d$ leads to a lower assistive value than $\hat{A}_d$. Utilizing the collected training set, we learn to critique, refine, and rank decompositions to improve assistive value. Figure \ref{fig:overview} presents the overview of our framework.

\subsection{Data Collection} \label{section:data_collection}

\paragraph{Data Preparation} For each problem $P$ and an initial solution $A$, we obtain $K$ different decomposed solutions $\{A_d^1, \cdots, A_d^K\}$ by sampling from various LMs with few-shot prompting. See more details about data preparation in Appendix \ref{appendix:data_preparation}.

\paragraph{Recruiting Human Annotators}\label{section:recruting} We recruit annotators from college students. We conduct a pre-survey about their background and divide them into two groups based on their programming skills. We recruit 11 expert annotators who can solve Leetcode hard-level problems. Especially, 6 of them are medalists in the National or International Olympiad in Informatics. We recruit 19 non-expert annotators who can solve Leetcode medium-level problems but hardly solve hard-level problems. We conduct a warm-up test to train annotators to use our experiment environment. We further verify annotators' programming skills based on their performance in the warm-up test.

\paragraph{Collecting Assistive Value}
Following our definition of assistive value in Equation \ref{equation:efficiency}, we collect assistive value labels of different decompositions in the actual human annotation process. Specifically, for each sampled decomposed solution $A^i_d$, we collect the following labels:
\begin{itemize}[leftmargin=12pt,itemsep=-1.5mm, topsep=1mm ]
\item \textbf{Assistive Value $\eta^i$:} We calculate the assistive value of $A_d^i$ following the definition described in Equation \ref{equation:efficiency}.
\item \textbf{Critique on decomposition $C^i$:} After repairing, we ask annotators to provide natural language critique $C^i$ on how decomposition assists or hinders their debugging.
\end{itemize}

\paragraph{Constructing Pair-wise Demonstrations}
We make $(A_d^j, A_d^i)$ a comparison pair if they meet two requirements: (1) There is a substantial difference in assistive value between $\eta^j$ and $\eta^i$. (2) The advantages in critique $C^i$ match the disadvantages in $C^j$\footnote{In our experiments, we manually inspect human critiques and perform matching, as we only need a few examples for in-context learning. See Appendix \ref{appendix:data_preparation} for more discussions.}, and hence $C^i$ and $C^j$ explain why $A_d^j$ leads to more efficient human repair than $A_d^i$.

\subsection{Models} \label{section:models}

Starting with a vanilla decomposition generated by naively prompting LMs, we predict human critiques on it with $\pi_\text{critic}$, refine the decomposition according to the critique with $\pi_\text{refine}$, and rank candidate decompositions to select the final high-\ourobjective output with $\pi_\text{rank}$. Figure \ref{fig:overview} presents the overview of our framework. Notably, we introduce critique as an integral step instead of directly generating refined decomposition since it provides enriched information for models to learn how decomposition can achieve improved assistive value. Additionally, it also enables controllable decomposition as humans can manually edit critiques (e.g., requiring specific decomposition on certain complex subtasks).

We next describe the detailed inputs and outputs of each model:
\vspace{-1mm}
\begin{itemize}[leftmargin=*,itemsep=-1.5mm, topsep=1mm ]
    \item \textbf{Critic Model $\pi_\text{critic}$}: It takes a problem $P$ and a decomposed solution $A_d$ as inputs, and outputs critique on how to improve $A_d$ for higher assistive value.
    \item \textbf{Refine Model $\pi_\text{refine}$}: It takes a problem $P$, a decomposed solution $A_d$, and a critique $C$ as inputs, and outputs a refined decomposed solution $\hat{A}_d$.
    \item \textbf{Ranking Model $\pi_\text{rank}$}: It takes a problem $P$ and two decomposed solutions $A_d^1$, $A_d^2$ as inputs, and outputs a ranking that predicts which decomposition leads to higher assistive value.
\end{itemize}

For model training, inspired by recent works showing that modern LMs can learn to critique and refine model outputs via in-context learning \cite{bai2022constitutional, sun2023principle}, we use this approach to learn our three models, where the collected training data $D_{train}=\{(\hat{A}_d,A_d,C,P)\}$ are formatted into in-context examples for each model. Example prompts are shown in Appendix \ref{appendix:prompt}.

At inference time,  we apply $\pi_\text{critic}$, followed by $\pi_\text{refine}$, to produce a refined decomposition set $\{\hat{A}_d\}$, and then use $\pi_\text{rank}$ to select a decomposition $A_d^*$ among them as the final decomposition.
\section{Experiments}

\begin{figure*}[t!]
\centering
\begin{minipage}[t]{0.4\linewidth}
\centering
\includegraphics[width=0.6\linewidth]{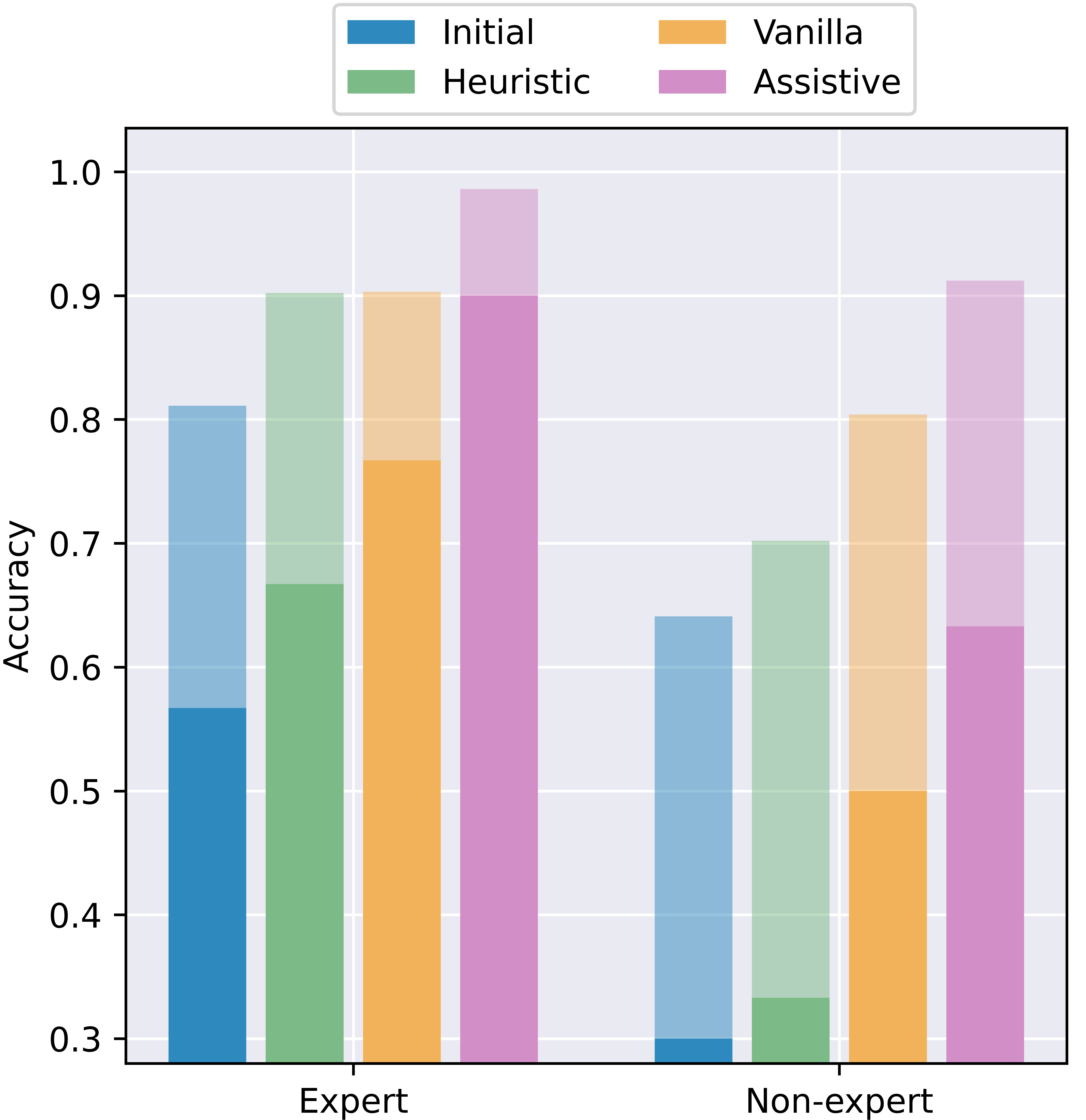}
\caption{Humans provide higher-quality labels with decomposition. Dark color denotes strict accuracy of human-repaired programs; light color denotes test case average accuracy.}
\label{fig:human_debugging_acc}
\end{minipage}
\hspace{.15in}
\begin{minipage}[t]{0.55\linewidth}
\centering
\includegraphics[width=\linewidth]{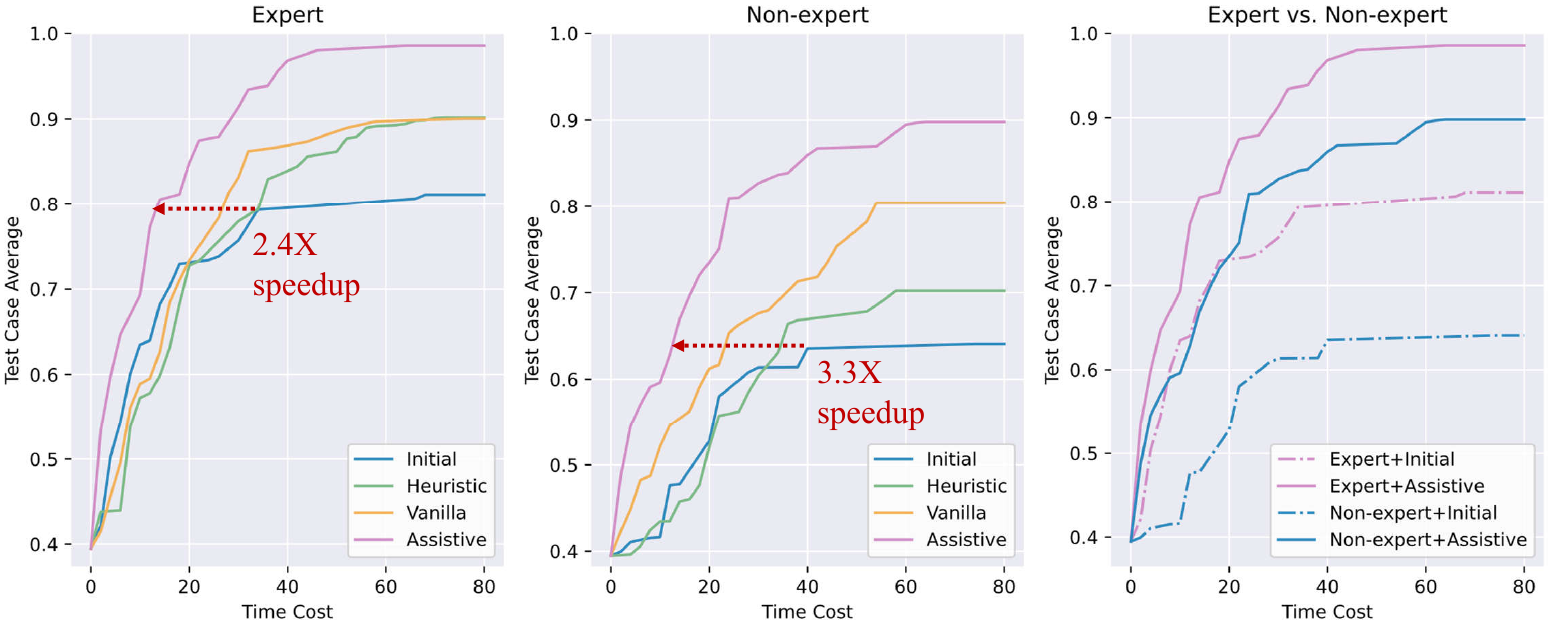}
\caption{Decomposition improves human efficiency. We plot the relationship between the human-repaired program's test case average accuracy and human time cost.}
\label{fig:human_debugging}
\end{minipage}
\end{figure*}

\subsection{Setup}
\paragraph{Benchmark} We conduct experiments with the problems from two widely adopted competition-level code generation benchmarks, namely APPS \cite{hendrycks2021measuring} and Code-Contests \cite{li2022competition}. For human evaluation, we filter those problems where the model-generated program directly passes all test cases and randomly sample 30 problems as the test data.

\paragraph{Metric} We measure the supervision quality in competitive programming by programs' passing rates on unit test cases. Specifically, we aggregate the program's performance on test cases with two metrics. \textbf{Test Case Average Accuracy} computes the average fraction of test cases passed among all the test cases. \textbf{Strict Accuracy} computes the average fraction of programs that pass all test cases.

\paragraph{Baselines} We consider three baselines:
\begin{itemize}[leftmargin=12pt,noitemsep,topsep=5pt]
\item \textbf{Initial}: It prompts a code language model $\mathcal{M}$ with the problem $P$ to generate a solution $A$ without explicit instructions for decomposition. 
\item \textbf{Heuristic Decomposition}: Inspired by McCabe's cyclomatic complexity \cite{mccabe1976complexity}, a widely adopted metric for measuring code complexity in software engineering, we implement a heuristic baseline to decompose a complex program into simpler pieces. Specifically, we decompose each if-statement as well as for or while loops into a separate function since these code structures contribute to higher cyclomatic complexity. We then generate post-hoc subtask descriptions using GPT-4.
\item \textbf{Vanilla Decomposition}: It prompts a code language model $\mathcal{M}$ to perform decomposition with basic author-crafted demonstrations without learning from human feedback.
\end{itemize}

\paragraph{Annotation Procedure}
We conduct experiments based on an internal Online Judge system. Problems are randomly assigned to each annotator while ensuring that they have not seen the assigned problem before (if they have, the problem will be reassigned) and never repeatedly debug the same problem. Next, given a competition-level problem, several exemplified public test cases, and a solution, labelers are required to perform debugging. See Appendix \ref{appendix:human_annotation} for more annotation details.

\paragraph{Implementation Details}
We use GPT-4 \cite{openai2023gpt4} as our default backbone model $\mathcal{M}$ for code generation and decomposition due to its superior in-context learning capabilities. To focus our evaluation on the impact of decomposition, we ensure the consistency between initial and decomposed solutions based on their outputs on test cases. If the consistency check fails, we reject the decomposition and retry within a sampling budget. The consistency check is applied in all baselines for fair comparisons. See Appendix \ref{appendix:consistency} for more implementation details.


\subsection{Assisting Human Supervision}

We evaluate the effectiveness of decomposition in aiding humans to repair programs and provide reliable supervision signals.

\subsubsection{Quantitative Analysis}

\paragraph{Supervising competitive programming is extremely challenging} In our experiments, highly experienced experts spend at least 68 minutes repairing 56.7\% of the model-generated programs, and non-experts spend at least 74 minutes repairing 30\% of the model-generated programs. These results indicate that competitive programming serves as a good testbed for studying scalable oversight.

\paragraph{Decomposition improves human efficiency} Figure \ref{fig:human_debugging} demonstrates that humans assisted with our decomposition model significantly outperform those without the assistance in terms of the speed to repair model-generated programs. For instance, the time required to collect repaired programs with 68\% test case accuracy from non-experts is reduced from 40 minutes to 12 minutes, achieving a 3.3$\times$ speedup. Paired $t$-tests confirm that these efficiency gains are statistically significant ($p<0.005$) for both experts and non-experts. These results indicate that our decomposition model can effectively improve the efficiency of human labelers.

\paragraph{Humans provide higher-quality labels with decomposition} Beyond efficiency, we evaluate how decomposition impacts the final quality of human labels. As shown in Figure \ref{fig:human_debugging_acc}, our decomposition model enables humans to repair more model-generated programs. For instance, decomposition improves the strict accuracy of repaired programs from 56.7\% to 90\% for experts and from 30.0\% to 63.3\% for non-experts. Paired $t$-tests confirm the significance ($p<0.01$) of these improvements. This verifies the effectiveness of decomposition in aiding humans to tackle complex tasks by breaking them down into more manageable subtasks. 

\paragraph{Decomposition enables non-experts to be comparable with experts}
Remarkably, when assisted with our decomposition model, non-experts achieve performance comparable to non-assisted experts in labeling efficiency and quality. This aligns with the overarching goal of scalable oversight, empowering human labelers to oversee models in superhuman tasks.

\begin{figure}[t!]
    \centering
    \includegraphics[width=0.75\linewidth]{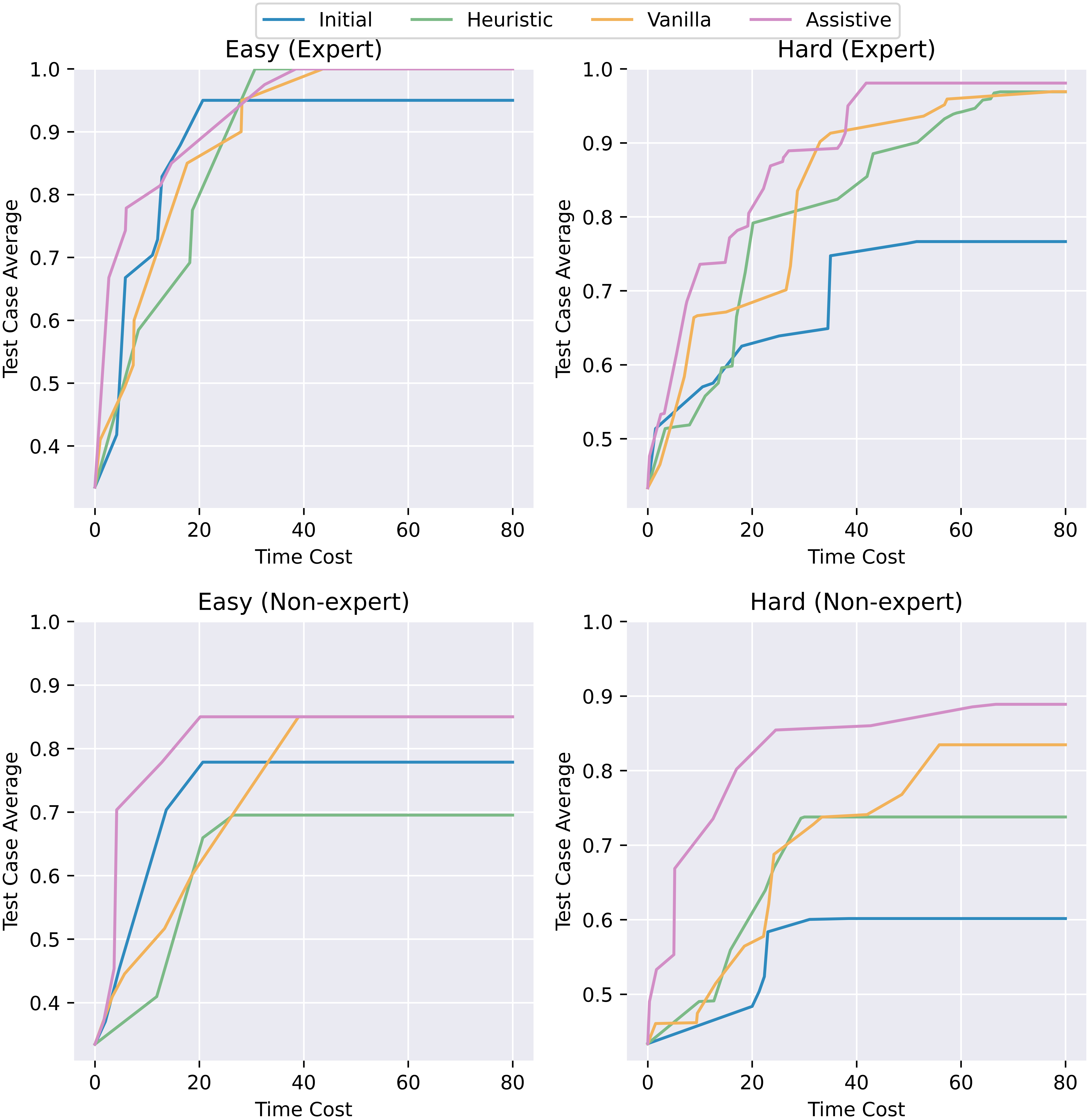}
    \caption{Decomposition brings more benefits to human labelers on hard problems.}
    \label{fig:human_debugging_group}
\end{figure}

\paragraph{Decomposition benefits more on hard problems} To further understand the impact of decomposition, we extract two subsets from our test data based on the time spent by non-experts to repair the model-generated programs: (1) Easy: problems that take less than 25 minutes. (2) Hard: problems that take more than 40 minutes. The results shown in Figure \ref{fig:human_debugging_group} reveal that easy problems benefit minimally from decomposition. However, when it comes to hard problems, decomposition significantly improves the labeling efficiency and final quality of human labelers. These results further underscore our decomposition's utility in complex tasks.

\paragraph{Humans prefer our decomposition model} 

\begin{figure}[t!]
    \centering
    \includegraphics[width=0.7\linewidth]{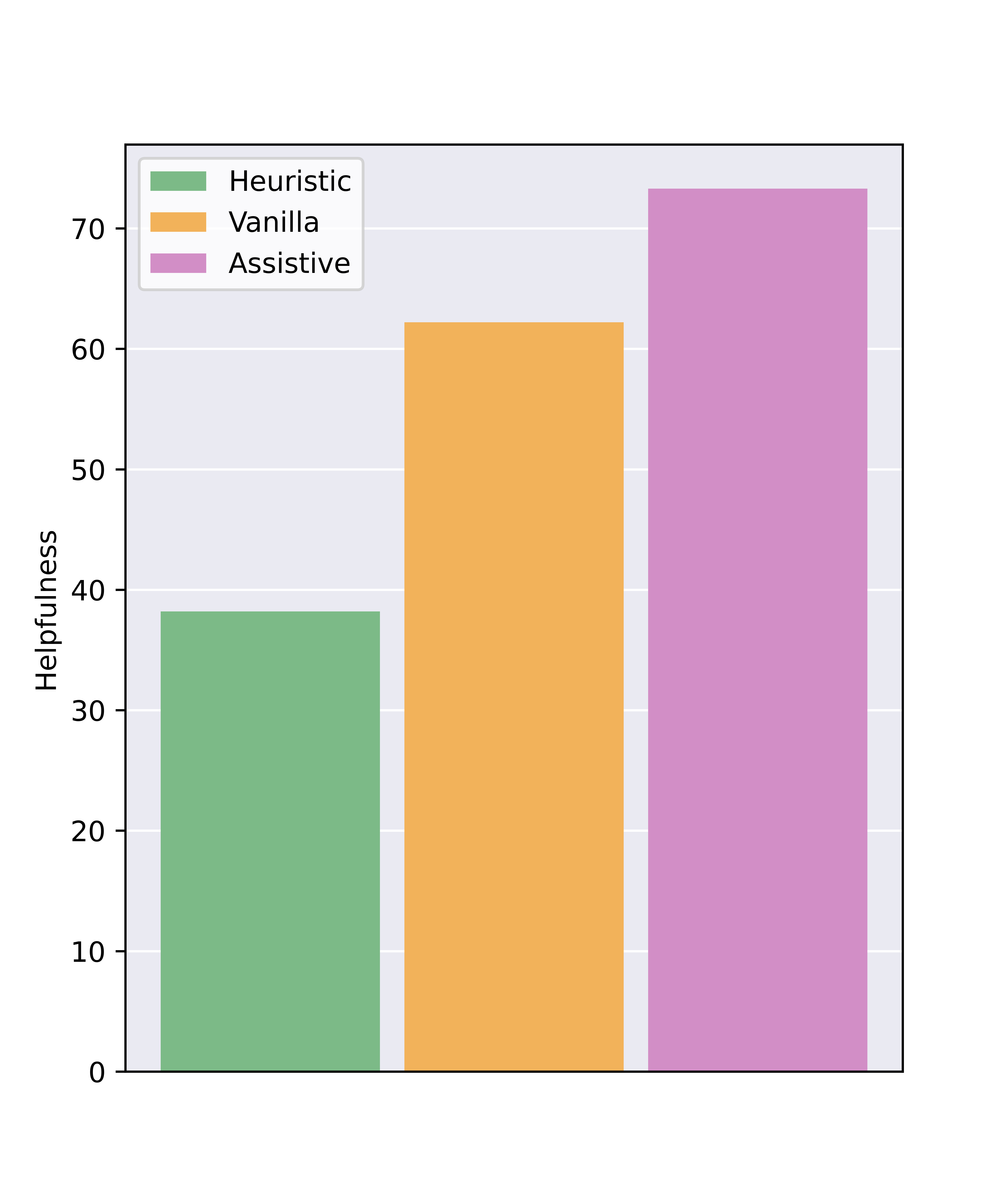}
    \caption{Helpfulness of decompositions evaluated by human labelers after repairing.}
    \label{fig:helpfulness}
\end{figure}

As shown in Figure \ref{fig:human_debugging_acc} and Figure \ref{fig:human_debugging}, our decomposition model surpasses other decomposition baselines in enhancing human labeling efficiency and quality. We also ask labelers to consider whether the decomposition is helpful for them to repair the program in the post-survey. As shown in Figure \ref{fig:helpfulness}, labelers rate our decomposition results substantially more helpful than the baselines'. These results verify that LMs can learn to produce high-\ourobjective decompositions by learning from human repair experiences.

\subsubsection{Qualitative Analysis} \label{sec:qualitative}
To further gain insight into what our decomposition model has learned from human feedback to enhance its utility, we analyze the collected human repairing traces alongside the corresponding critique, from which we draw the following qualitative observations.

\paragraph{Assisting humans by highlighting boundary conditions} We observe that decomposing complex programs around boundary conditions effectively aids humans. As boundary conditions typically involve more straightforward logic than the core problem, humans can more easily understand and repair them. For instance, in one training demonstration (Figure \ref{fig:case_boundary_condition_1}), \inlinecode{check\_validity} highlights all boundary conditions, earning positive human feedback for enabling them to quickly identify that problems are located in \inlinecode{calculate\_minimum\_moves}. This principle is learned by our decomposition model, as evidenced in Figure \ref{fig:case_boundary_condition_2}, where it highlights two boundary conditions in \inlinecode{check\_zero\_piles} and \inlinecode{check\_same\_piles}, leading humans to quickly realize the missing third boundary condition.

\paragraph{Assisting humans by creating simpler subtasks} Decomposition reduces human workload by breaking down the initial solution into multiple simpler pieces, each corresponding to a specific subtask. For instance, in one training demonstration (Figure \ref{fig:case_simple_subtask_1}), \inlinecode{find\_cycle} simplifies the complex \inlinecode{find\_cycles}, and \inlinecode{generate\_permutation} helps humans locate the actual bugs in a simple subtask. Similarly, in Figure \ref{fig:case_simple_subtask_2}, our decomposition model creates a simple task \inlinecode{calculate\_participants} which effectively isolates bugs. In addition, our decomposition model learns to decompose code pieces that humans typically struggle with (e.g., nested loops, binary search, and dynamic programming), as exemplified in Figure \ref{fig:case_simple_subtask_1} and Figure \ref{fig:case_high_level_1}.

\paragraph{Assisting humans by presenting clear high-level logic} Decomposition's ability to offer clear high-level logic can accelerate comprehension and bug identification before delving into low-level details, as demonstrated in Figure \ref{fig:case_high_level_1}. Next, Figure \ref{fig:case_high_level_2} illustrates our decomposition model's integration of this principle by creating two subtasks \inlinecode{toggle\_doors} and \inlinecode{toggle\_single\_door}. This high-level logic indicates that the current solution addresses each door independently, thereby enabling humans to locate bugs directly since doors are interrelated in the problem context.

\subsection{Assisting AI Supervision}

With the verified effectiveness of decomposition in assisting human supervision, we further explore its potential to assist AI supervision. We suspect that some patterns of decompositions shown in Section \ref{sec:qualitative} (e.g., creating simpler subtasks and presenting clear high-level logic) can also help AI to evaluate (i.e., discriminate programs' correctness) and repair programs. We achieve AI supervision by prompting a code language model $\mathcal{M}$ to perform discrimination and repair \cite{saunders2022self,madaan2023self,chen2023teaching}.

\begin{table}[t!]
\centering
\resizebox{\linewidth}{!}
{
    \begin{tabular}{lccc}
    \toprule
    \multirow{2}{*}{\textbf{Program}} & \textbf{Discrimination} & \multicolumn{2}{c}{\textbf{Repair}} \\
    & \textbf{Acc} & \textbf{Acc (Strict)} & \textbf{Acc (Test Case)} \\
    \midrule
    \multicolumn{4}{c}{{\cellcolor[gray]{.95}}\textit{APPS}}\\
    \midrule
    \textbf{Initial} & 10.2 & 18.3~\ctext[RGB]{255,251,204}{(+0.0)} & 41.5~\ctext[RGB]{251,235,235}{(-0.3)}\\
    \textbf{Heuristic Decomp} & 30.0 & 16.7~\ctext[RGB]{244,204,204}{(-1.6)} & 39.2~\ctext[RGB]{244,204,204}{(-2.6)} \\
    \textbf{Vanilla Decomp} & 32.7 & 19.2~\ctext[RGB]{223,255,223}{(+0.9)} & 42.7~\ctext[RGB]{223,255,223}{(+0.9)}\\
    \midrule
    \textbf{Assistive Decomp} & \textbf{42.9} & \textbf{21.7}~\ctext[RGB]{191,255,190}{(\textbf{+2.9})} &  \textbf{47.4}~\ctext[RGB]{191,255,190}{(\textbf{+5.6})} \\
    \midrule
    \multicolumn{4}{c}{{\cellcolor[gray]{.95}}\textit{Code-Contests}}\\
    \midrule
    \textbf{Initial}& 26.7 & 6.3~\ctext[RGB]{255,251,204}{(+0.0)} & 20.1~\ctext[RGB]{242,255,210}{(+1.3)}\\
    \textbf{Heuristic Decomp} & 46.7 & 7.3~\ctext[RGB]{242,255,210}{(+1.0)} & 19.3~\ctext[RGB]{255,251,204}{(+0.5)} \\
    \textbf{Vanilla Decomp} & 53.1 & 8.3~\ctext[RGB]{223,255,223}{(+2.0)} & 21.4~\ctext[RGB]{223,255,223}{(+2.6)} \\
    \midrule
    \textbf{Assistive Decomp} & \textbf{62.5} & \textbf{12.5}~\ctext[RGB]{191,255,190}{(\textbf{+6.2})} &  \textbf{26.3}~\ctext[RGB]{191,255,190}{(+\textbf{7.5})} \\
    \bottomrule
    \end{tabular}
}
\caption{Decomposition aids AI systems to discriminate and repair programs, improving discrimination accuracy and repaired solutions' accuracy. Parenthetical values denote the accuracy variation from non-repair programs.}
\label{tab:AI_debugging}
\end{table}

\begin{figure}[t!]
    \centering
    \includegraphics[width=0.7\linewidth]{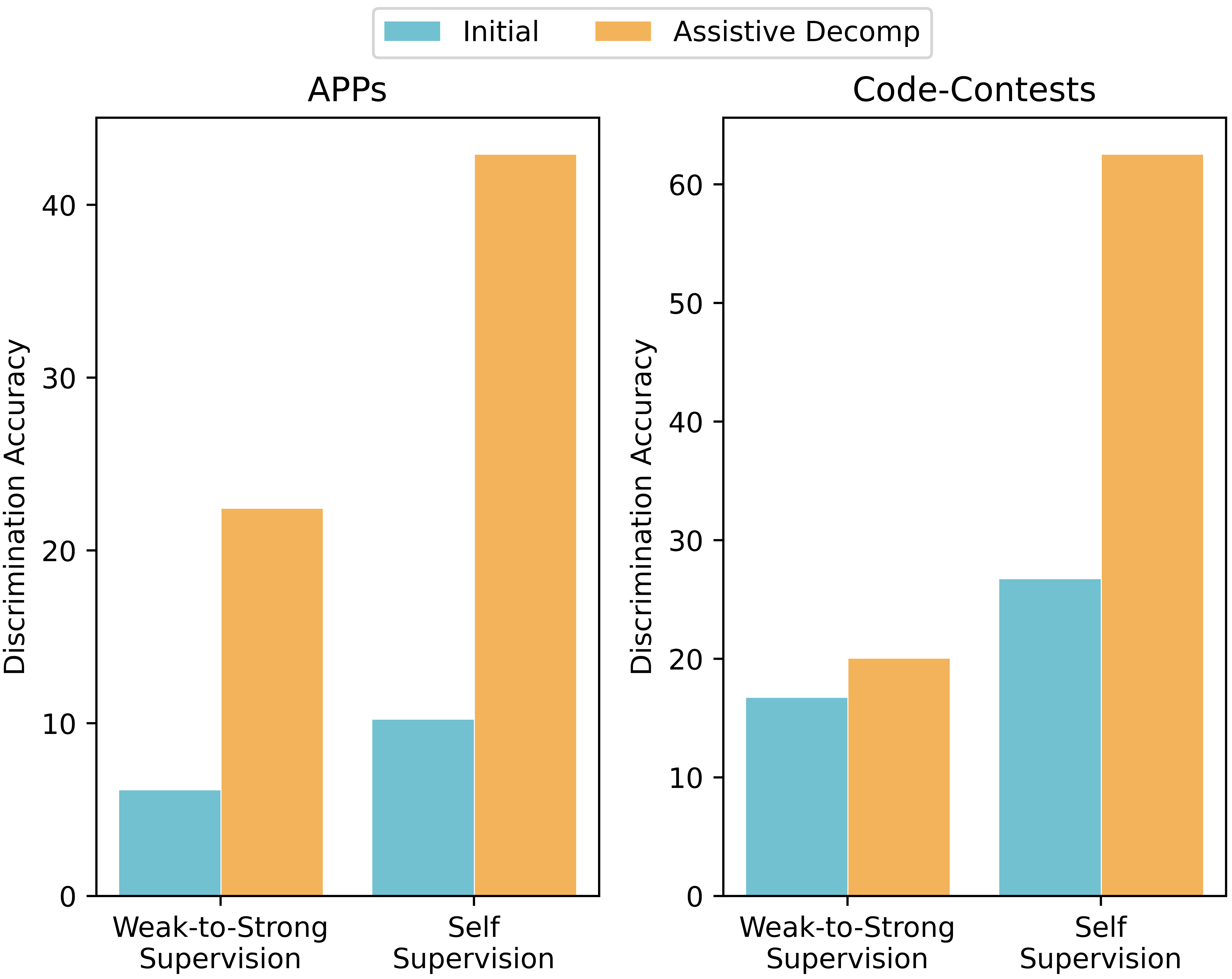}
    \caption{Decomposition aids AI systems to provide self-supervision, where GPT-4 discriminates its own outputs, and weak-to-strong supervision, where GPT-3.5-turbo discriminates GPT-4-generated programs.}
    \label{fig:ai_weak_to_strong}
\end{figure}

\paragraph{Decomposition facilitates accurate self-supervision} \label{section:assist_AI}
We first investigate decomposition's impact on self-supervision, where we use GPT-4 to supervise itself. Results in Table \ref{tab:AI_debugging} reveal that GPT-4 struggles to discriminate and repair programs generated by itself, echoing recent findings in LLMs' self-correction abilities \cite{huang2023large}. However, decomposing the complex programs with our model improves both discrimination accuracy and repaired solution quality by a large margin. For instance, discrimination accuracy improves from 10.2\% to 42.9\% on APPS and from 26.7\% to 62.5\% on Code-Contests. 

\paragraph{Decomposition facilitates accurate weak-to-strong supervision}
Having verified the effectiveness of decomposition in aiding non-expert human labelers, we now investigate its potential to assist weak models in supervising strong models. We evaluate GPT-3.5-turbo's ability to discriminate the programs generated by GPT-4. The results in Figure \ref{fig:ai_weak_to_strong} reveal that our method leads to more accurate weak-to-strong AI supervision. For instance, it improves GPT-3.5-turbo's discrimination accuracy from 6.1\% to 22.4\% on APPS, surpassing the non-assisted accuracy of GPT-4 (10.2\%).

\section{Analysis}
\begin{table}[t!]
    \centering
    \small
    \begin{tabular}{lc}
    \toprule
    \textbf{Method} & \textbf{Accuracy}\\
    \midrule
    \textbf{Intuitive Human Preference} & 49.5\\
    \midrule
    \textbf{Cyclomatic Complexity} & 48.4\\
    \midrule
    \textbf{GPT-3.5-turbo Zero-shot} & 54.3 \\
    \textbf{GPT-3.5-turbo Few-shot} & 62.5 \\
    \midrule
    \textbf{GPT-4 Zero-shot} & 66.7 \\
    \textbf{GPT-4 Few-shot ($\pi_\text{rank}$)} & \textbf{78.1} \\  
    \bottomrule
    \end{tabular}
    \caption{Accuracy of the rank model in predicting the assistive value of decompositions in actual human repair processes.}
    \label{tab:rank_agreement}
\end{table}

\subsection{Validity of the Rank Model} \label{section:validity_rank}

We evaluate the effectiveness of our ranking model $\pi_\text{rank}$ in predicting the assistive value of decompositions. We construct paired decompositions annotated with real assistive value labels as the test data. The results shown in Table \ref{tab:rank_agreement} indicate that LLMs can learn to select higher-\ourobjective decomposition for assisting humans via in-context learning, surpassing prior heuristic or zero-shot baselines. Notably, we ask humans to provide preferences on paired decompositions without repairing them, which is shown to align poorly with their actual assistive value for human repair. These results highlight the importance of measuring assistive value based on actual human supervision experiences. 

See Appendix \ref{appendix:rank_model} for more details about these ranking baselines.

\subsection{Ablations on the Critic Model}
We perform an ablation to explore the impact of our critic model $\pi_\text{critic}$. In the ablated version, we directly generate refined decompositions without incorporating predicted critiques. We use $\pi_\text{rank}$ as a proxy to evaluate the assistive value of decompositions due to its reasonable performance (Section \ref{section:validity_rank}). The win rates between models with and without $\pi_\text{critic}$ are 58.8\% and 35.3\%, respectively. These results indicate the effectiveness of learning from informative natural language human feedback.

\subsection{Distilled Decomposition Model}
In this section, we aim to study whether moderate-size open-source LLMs can also learn to generate better decompositions to assist human or AI supervision while reducing API costs and enhancing reproducibility. We hence distill the knowledge of assistive decomposition from proprietary LLMs into an in-house model $\pi_\theta$. Specifically, we create supervised data $D=\{(P, A, A_d^*)\}$ with LLMs as illustrated in Section \ref{section:models}, and optimizing $\pi_\theta$ over $D$ via the standard MLE loss:

$$
    \mathcal{L}_{MLE} = -\sum_{t=1}^{|A_d^*|}\text{log}\pi_\theta(A^*_{d_t}|A^*_{d_{<t}}, A, P) 
$$
where $P$, $A$, and $A_d^*$ denote the problem, the initial model-generated solution, and the decomposed solution, respectively.

We evaluate the performance of the Code-LLaMA-based \cite{roziere2023code} distilled decomposition model in aiding human and AI supervision, finding it outperforms the vanilla decomposition based on Code-LLaMA and even GPT-4. These results reveal the effectiveness of distilling the knowledge of assistive decomposition from GPT-4 to Code-LLaMA. See Appendix \ref{appendix:distill} for detailed results.

\section{Related Work}

\subsection{Code Generation} Language models have achieved impressive performance in generating simple code pieces \cite{chen2021evaluating, austin2021program} by learning from human demonstrations \cite{li2023starcoder}, and are being used to generate increasingly complex programs \cite{hendrycks2021measuring,li2022competition}. In this context, our paper studies how to aid humans in efficiently providing reliable supervision to train models that don't write buggy programs in complex and high-impact scenarios.

\subsection{AI-assisted Programming} Prior works have explored various forms to assist human programmers, including code completion \cite{openai2021copilot}, bug location \cite{xie2016revisit}, and program repair \cite{joshi2023repair}. 
However, these methods rely on assistant models' capacity for accurate code generation or repair---a requirement unmet in our setup, where models face challenges in performing either task. We thus explore an alternative form of assistance via task decomposition, which alleviates the burden on assistant models.

\subsection{Task Decomposition}
Task decomposition has been extensively studied for tackling complex tasks, primarily focusing on enhancing model performance \cite{khot2022decomposed, dua2022successive,yao2022react} and enabling efficient searching over the subtask space \cite{zelikman2023parsel}. In contrast, our paper studies decomposition from a human-centric perspective, aiming to facilitate human supervision over complex tasks, akin to \citet{wu2021recursively}'s work in aiding human supervision over book summarization. Concerning the implementation of decomposition, prior works mainly rely on fixed heuristic rules (e.g., decomposing each $N$-line of codes) or author-crafted demonstrations \cite{wu2021recursively,khot2022decomposed, zelikman2023parsel}. In this paper, we take the first step towards advanced decomposition by learning to improve a novel objective: assistive value, which exactly measures the feasibility and speed of human repair in the actual annotation process.

\subsection{Learning from Human Feedback}
Learning from human feedback has been widely adopted in recent works for developing human-friendly general language models \cite{ouyang2022training}. Our work can also be understood as introducing this concept to the development of assistant models. Specifically, our introduced objective---assistive value--could also be introduced as a unique type of human feedback, which is gathered from actual human problem-solving experiences. Therefore, the collected human feedback could be more aligned with the ultimate goal of AI assistants: enhancing human problem-solving performance.

\subsection{Scalable Oversight} 
Advancements in LLMs have intensified the need for scalable, reliable human oversight on complex tasks that reach or even exceed the capabilities of human experts. Addressing this, prior research has explored various assistance methods, such as self-critiquing \cite{saunders2022self}, AI debate \cite{parrish2022two}, using simple and informative examples \cite{zhong2023non}, and decomposition \cite{christiano2018supervising, wu2021recursively}. 
However, previous works overlook the feedback from actual assisted humans, potentially reducing helpfulness for humans \cite{xie2016revisit,parrish2022two}. 
To bridge this gap, we propose to learn assistance models from human feedback.

\section{Conclusion}
This paper focuses on assisting humans in supervising LMs on complex problems by automated task decomposition. We introduce a novel objective for learning task decomposition: assistive value (\ourobjective), which measures the feasibility and speed of humans to repair a decomposed solution. We collect a dataset of decompositions, measure their \ourobjective, and ask human annotators to provide a natural language critique on what makes a decomposition (not) helpful. We then learn to critique, refine, and rank decompositions to generate high-\ourobjective decompositions. Experiment results demonstrate that our method can effectively assist humans in providing higher-quality supervision with significantly less time. Notably, our method assists non-experts in matching unassisted experts. Overall, we show that even when LMs cannot solve the problem themselves, they can still learn to assist humans by learning from human repair experiences. These results highlight the potential of learning-based methods to assist humans with scalable oversight.

\section*{Limitations and Future Work}
One limitation of our work is that our non-expert participants, who could solve 30\% of the problems while spending 74 minutes, are still more skilled than novice programmers (e.g., programmers who can only solve Leetcode easy-level problems) or non-programmers. Nonetheless, we demonstrate the effectiveness of our decomposition model, which successfully enables assisted non-experts to be comparable with non-assisted experts. We leave the investigation of leveraging decomposition to assist novice programmers or even non-programmers as future work.

Another limitation of our paper is that we did not conduct experiments on misleading benchmarks, which mainly consist of subtle errors that humans tend to overlook. The effectiveness of decomposition in aiding humans in detecting and fixing these subtle errors is worth studying in future work.

Moreover, we only conduct experiments based on in-context learning and supervised learning. Given the reasonable performance of the rank model (Section \ref{section:validity_rank}), fine-tuning the decomposition model with reinforcement learning is worth studying in the future.

Finally, our introduced objective \ourobjective, which highlights the possibility of learning an effective assistance model from human problem-solving experiences, can be further extended to other assistance forms beyond decomposition. For example, recent works explore using model-generated explanations to assist humans in complex decision-making tasks (e.g., complex question answering \cite{rein2023gpqa}). However, the practical benefits of model explanations are mixed as they may introduce additional human workload for understanding the explanation or even mislead humans to incorrect results \cite{parrish2022two}. Therefore, future works can explore what kind of explanations can better assist humans by measuring the \ourobjective of different explanations and learning to improve \ourobjective. 

\section*{Ethics Statement}
We recruit annotators mainly from college students. We inform annotators in advance how their annotation data will be collected and used. We pay them 30 USD per hour, which is higher than the average wage of the local residents.

\section*{Acknowledgements}

This work was supported by the National Science Foundation for Distinguished Young Scholars (with No. 62125604) and the NSFC projects (with No. 62306160). This work was also supported by China National Postdoctoral Program for Innovative Talents (No. BX20230194) and China Postdoctoral Science Foundation (No. 2023M731952). RZ is supported by Simons Foundation fund, chartstring 71815-13090-44–PSJST.

\bibliography{custom}

\appendix


\section{Implementation Details}

\subsection{Consistency of Decomposition} \label{appendix:consistency}
To ensure consistency between initial and decomposed solutions, we set the max retry time $N=8$. If the consistency check fails in the end, we directly use the initial solution as the final output. Empirically, we find the consistency ratio of the decomposition generated by few-shot prompting with Code-LLaMA-13B, GPT-3.5-turbo, and GPT-4 is 92.9\%, 97.6\%, and 98.8\%, respectively.

\subsection{Data Collection}  \label{appendix:data_preparation}
\paragraph{Data Preparation} We use vanilla decomposition to sample $K=5$ different decomposed solutions from various code generation models, including Code-LLaMA-13B, GPT-3.5-turbo, and GPT-4. We use top-p sampling with a default temperature 0.5. 

\paragraph{Constructing Pair-wise Demonstrations}
For constructing pair-wise decompositions, since we adopt a few-shot learning setup in our experiments, we manually inspect the collected human critiques and perform matching. And we finally collect five pair-wise decompositions as the training data for in-context learning. 

Considering that matching critiques based on semantic similarity is not challenging for modern LMs, we further explore automatically matching decompositions based on language models. Specifically, given two decomposed solutions with the corresponding human critiques: $(A_d^1,C^1)$, $(A_d^2,C^2)$, we prompt GPT-3.5-turbo to evaluate if $C^1$ matches $C^2$. We construct a test set that contains 50 decomposition pairs for evaluating the automatic matching performance, where the golden labels are derived from manual matching results. We find that GPT-3.5-turbo achieves a perfect accuracy of 100\%. These results indicate that it is promising to construct pair-wise demonstrations automatically.

\subsection{Benchmark}
We obtain competition-level problems from Code-Contests and APPS.  We use the whole test set of Code-Contests that consists of 96 problems and randomly sample 120 problems from the competition-level subset of APPs. When conducting human experiments, we first filter those problems where the generated solution directly passes all the hidden test cases and randomly sample 30 problems as the test data.

\section{Additional Results}

\subsection{Comparing Human Propose with Human Repair}
\begin{figure}[t!]
    \centering
    \includegraphics[width=0.9\linewidth]{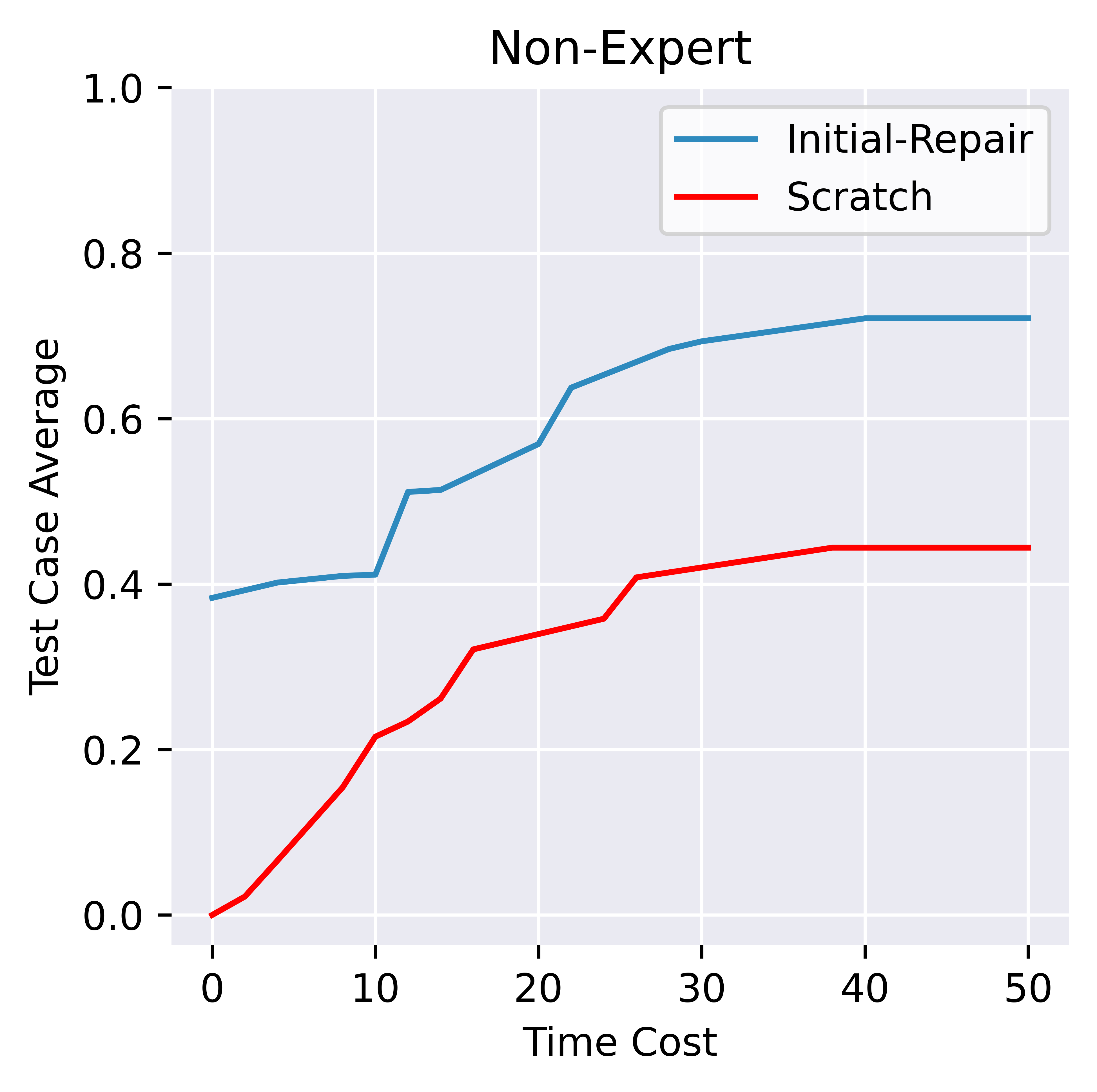}
    \caption{Comparing the performance of human labelers when asking them to write programs from scratch or repairing model-generated programs.}
    \label{fig:scratch}
\end{figure}

We adopt a propose-and-repair pipeline to obtain labels for code generation. Instead of burdening humans with creating programs from scratch, we leverage the coding capabilities of modern LLMs to propose initial programs and then ask humans to repair them. This pipeline is motivated by the recognition that modern LLMs can effectively solve a considerable portion of test cases, making them a practical starting point for reducing the human workload and streamlining the data collection process \cite{vaithilingam2022expectation}. 

We also empirically verify the effectiveness of this pipeline by comparing it with a baseline where human labelers are required to write programs from scratch. We construct two test sets for competitive programming, consisting of 10 and 18 problems that code generation models (e.g., GPT-4 in our experiments) succeed or fail to solve, respectively. We then study human labelers' efficiency on these two test sets. For problems that models can directly solve, non-expert humans also achieve 100\% accuracy while still spending on average 24.8 minutes. For problems that models fail to solve, the results are shown in Figure \ref{fig:scratch}. We can see that human labelers can also more easily collect a high-quality program label based on an initial model-generated program, yielding higher labeling quality and efficiency.

\subsection{Directly Generating Decomposed Solutions} \label{appendix:direct_decomposition}
In our experiments, we follow a two-stage framework to first generate an initial solution and then decompose it. We choose not to generate decomposed solutions for these two reasons. First, the two-stage framework enables us to focus on the impact of decomposition. Otherwise, directly generating decomposed solutions may introduce differences in solution contents, which can also impact human repair experiences. Second, we observe that code generation models might perform worse when prompted to generate decomposed solutions directly. Table \ref{tab:directy_prompting} presents the results, from which we can observe a substantial decrease in accuracy when directly prompting GPT-4 to generate decomposed solutions. We conjecture this might be due to the fact that most codes in these language models' pre-training corpus are not well modularized and decomposed.

\begin{table}[t!]
    \centering
    \resizebox{\linewidth}{!}
    {
        \begin{tabular}{lcc}
        \toprule
        \textbf{Prompt} & \textbf{Strict Accuracy} & \textbf{Test Case Average}\\
        \midrule
        Non-decomposed & \textbf{18.3} & \textbf{41.5} \\
        Decomposed & 16.7 & 37.4 \\
        \bottomrule
        \end{tabular}
    }
    \caption{Prompting GPT-4 to generate non-decomposed solutions and decomposed solutions. We evaluate the accuracy of the generated solutions on APPS.}
    \label{tab:directy_prompting}
\end{table}

\subsection{Evaluating Code Decomposition with Software Engineering Metrics}
We adopt the following four metrics to evaluate decomposition, which are widely adopted in software engineering:
\begin{itemize}[leftmargin=*,itemsep=-1.5mm, topsep=1mm ]
    \item Func Number: It calculates the average number of functions (i.e., subtasks)
    \item Avg Complexity: It first calculates the average cyclomatic complexity among all pieces in a single program and then takes an average across all programs.
    \item Max Complexity: It first calculates the maximum cyclomatic complexity among all pieces in a single program and then takes an average across all programs.
    \item Global Max Complexity: It calculates the maximum cyclomatic complexity among all pieces across all programs.
\end{itemize}

From the results shown in Table \ref{tab:code_stats}, we can see that by learning from human feedback, our decomposition model produces more subtasks and lower complexity than the vanilla decomposition baseline. In addition, while heuristic decomposition results in the largest number of subtasks and the lowest cyclomatic complexity, it does not effectively assist humans in practice. These results indicate that these naive objectives for code decomposition (e.g., function number, cyclomatic complexity) are not well aligned with the assistive value in the actual human supervision process.

\begin{table}[t!]
    \centering
    \resizebox{\linewidth}{!}
    {
        \begin{tabular}{lccccc}
        \toprule
        \multirow{2}*{\textbf{Program}} & \multirow{2}*{\textbf{Func Number}} & \multicolumn{3}{c}{\textbf{Complexity}} \\
        & & \textbf{Avg} & \textbf{Max} & \textbf{Global Max} \\
        \midrule
        \multicolumn{5}{c}{\cellcolor[gray]{.95}\textit{APPS}}\\
        \midrule
        Initial & 0.4 & 5.5 & 5.7 & 17.0\\
        Heuristic Decomp & 4.5 & 2.0 & 2.3 & 4.0\\
        Vanilla Decomp & 2.8 & 2.5 & 4.0 & 12.0\\
        Human-Centric Decomp & 3.1 & 2.4 & 3.8 & 10.0\\
        \midrule
        \multicolumn{5}{c}{\cellcolor[gray]{.95}\textit{Code-Contests}}\\
        \midrule
        Initial & 0.4 & 5.1 & 5.7 & 18.0\\
        Heuristic Decomp & 5.2 & 2.0 & 2.3 & 5.0\\
        Vanilla Decomp & 2.9 & 2.7 & 4.2 & 12.0\\
        Human-Centric Decomp & 4.3 & 2.2 & 3.7 & 10.0\\
        \bottomrule
        \end{tabular}
    }
    \caption{Statistics of initial programs and decomposed programs. ``Func number'' denotes the average number of functions (i.e., subtasks), ``Complexity'' denotes the value of cyclomatic complexity. }
    \label{tab:code_stats}
\end{table}

\subsection{Analysis of the Rank Model} \label{appendix:rank_model}
Given a question $Q$, and a pair of decomposed solutions $A^1_d$ and $A^2_d$, we adopt the following four methods to predict which decomposed solution leads to higher assistive value:
\begin{itemize}[leftmargin=*,itemsep=-1.5mm, topsep=1mm ]
    \item \textbf{Intuitive Human Preference}: ask human programmers to give an intuitive preference for which decomposition can lead to higher human repair performance (i.e., higher assistive value). For each decomposition pair, we ask three annotators to give a preference. We adopt majority voting to make final decisions among three annotators.
    \item \textbf{Cyclomatic Complexity}: consider the solution with a lower cyclomatic complexity as the better one.
    \item \textbf{Zero-shot}: prompt a code language model $\mathcal{M}$ to select a more effective decomposition in a zero-shot setting.
    \item \textbf{Few-shot ($\pi_\text{rank}$)}: prompt a code language model $\mathcal{M}$ to select a more effective decomposition in a few-shot setting, where the few-shot demonstrations are collected from the feedback of real human labelers as illustrated in Section \ref{section:method}.
\end{itemize}

\subsection{Analysis of the Distilled Decomposition Model} \label{appendix:distill}
We compare the distilled assistive decomposition model based on Code-LLaMA-13B with the vanilla decomposition based on Code-LLaMA-13B and GPT-4, as well as its GPT-4-based teacher model. We conduct experiments to assist both human supervision and AI supervision.

\paragraph{Assisting Human Supervision}
\begin{figure}[t!]
    \centering
    \includegraphics[width=0.9\linewidth]{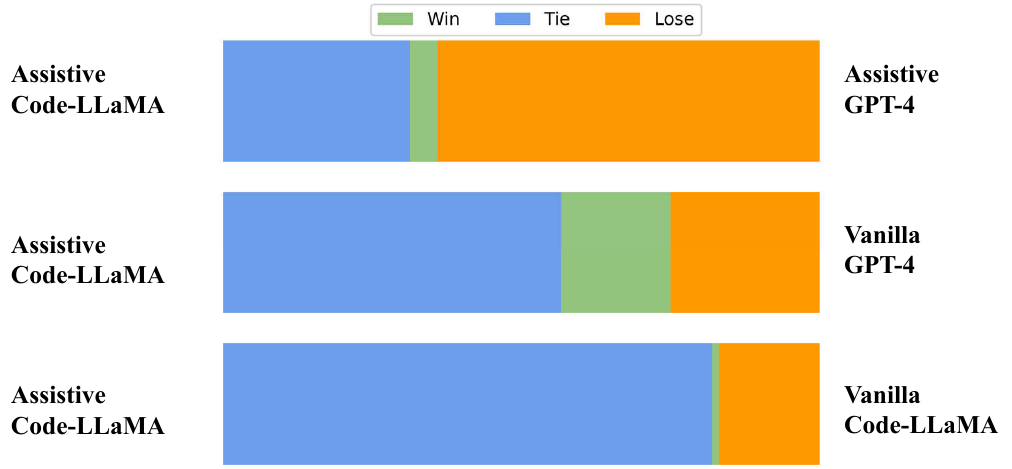}
    \caption{Comparison of the distilled assistive Code-LLaMA model against three other decomposition models. We use our rank model as a proxy evaluator.}
\label{fig:ai_debugging_distill_human}
\end{figure}
\begin{figure}[t!]
    \centering
    \includegraphics[width=0.9\linewidth]{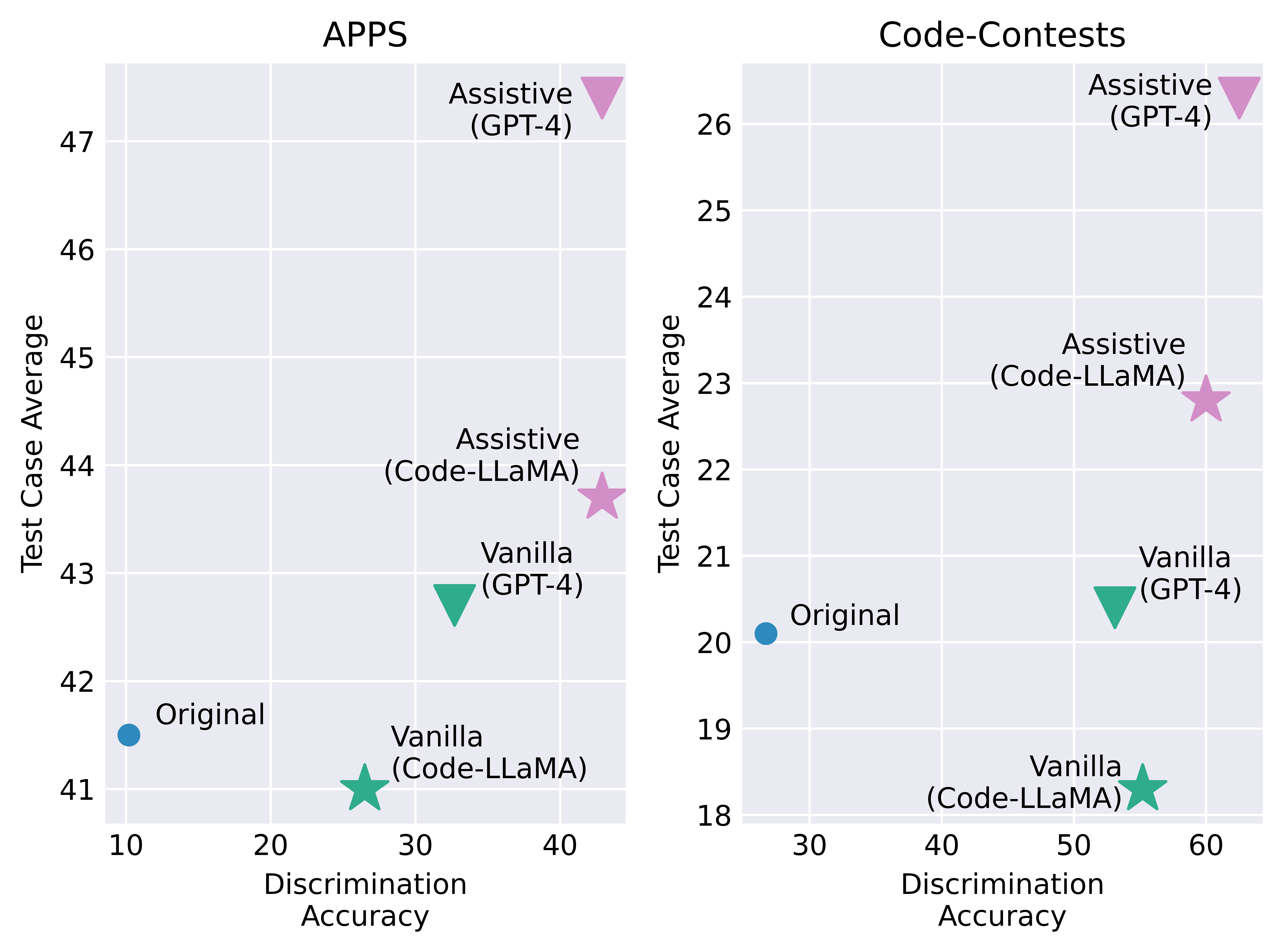}
    \caption{Assisting AI supervision with the distilled decomposition model.}
    \label{fig:ai_debugging_distill}
\end{figure}

Based on the reasonable accuracy of our rank model in predicting the actual assistive value of decompositions (Section \ref{section:validity_rank}), we use the prediction of our rank model as a proxy to evaluate the distilled decomposition model. The results shown in Figure \ref{fig:ai_debugging_distill_human} demonstrate that the distilled assistive Code-LLaMA model substantially outperforms vanilla Code-LLaMA and moderately outperforms vanilla GPT-4, thanks to the internalized high-\ourobjective decomposition knowledge in the generated data from its teacher model. 

\paragraph{Assisting AI Supervision}
As illustrated in Section \ref{section:assist_AI}, we evaluate the distilled decomposition model's performance in assisting AI systems in providing two forms of supervision: discrimination and repair. As shown in Figure \ref{fig:ai_debugging_distill}, the distilled decomposition model leads to more accurate AI supervision than the vanilla Code-LLaMA-13B model and even the vanilla GPT-4.

\section{Additional Human Annotation Details} \label{appendix:human_annotation}
\label{sec:appendix}

\paragraph{Labeler selection} Our labelers are mainly hired from college students. To ensure the honesty of labelers, we track their debugging trajectories and filter those who plagiarize online golden solutions. As for group slicing, besides conducting a pre-survey to collect their self-evaluation on the programming level, we further examine their level based on their performance during the warm-up test. 

\paragraph{Labeling instruction} We present the summary of our labeling instructions in Table \ref{tab:instruction}.

\begin{table*}[h]
    \centering
    \begin{tabular}{p{\textwidth}}
    \toprule
    \textbf{Instruction:}\\
    You are given an algorithmic coding problem and a model-generated solution. Your job is to debug the solution and improve its accuracy. \\\\
    During debugging, you can actively submit your code and run your custom test cases. If the solution has a modular structure with multiple subfunctions, leverage it to accelerate your debugging. For example, you can check the presented high-level logic before inspecting low-level implementations, and you can perform function-level debugging.
    \\\\
    There are two criteria to stop debugging: 
    (1) The repaired solution passes all hidden test cases.
    (2) The debugging time already exceeds 30 minutes. This timeframe is imposed to avoid endless debugging.\\
    \midrule
    \textbf{Survey:}\\
    Review your debugging process and answer the following questions.
    \vspace{-2mm} 
    \begin{itemize}
    \itemsep-1mm 
        \item Fixed Bugs: What bugs have you fixed during debugging?
        \item Critique on Decomposition: how does the current decomposition impede or improve your debugging efficiency?
        \item Other Assistance Forms: what other assistance forms do you need based on your debugging experience?
        \vspace{-4mm} 
    \end{itemize}
    \\
    \bottomrule 
    \end{tabular}
    \caption{Summary of our labeling instruction and survey.}
    \label{tab:instruction}
\end{table*}

\begin{table*}
    
\end{table*}

\paragraph{Interface}
In Figure \ref{fig:interface}, we present screenshots of our interface. 

\begin{figure*}[h]
     \centering
     \begin{subfigure}[b]{1.0\textwidth}
         \centering
         \includegraphics[width=\textwidth]{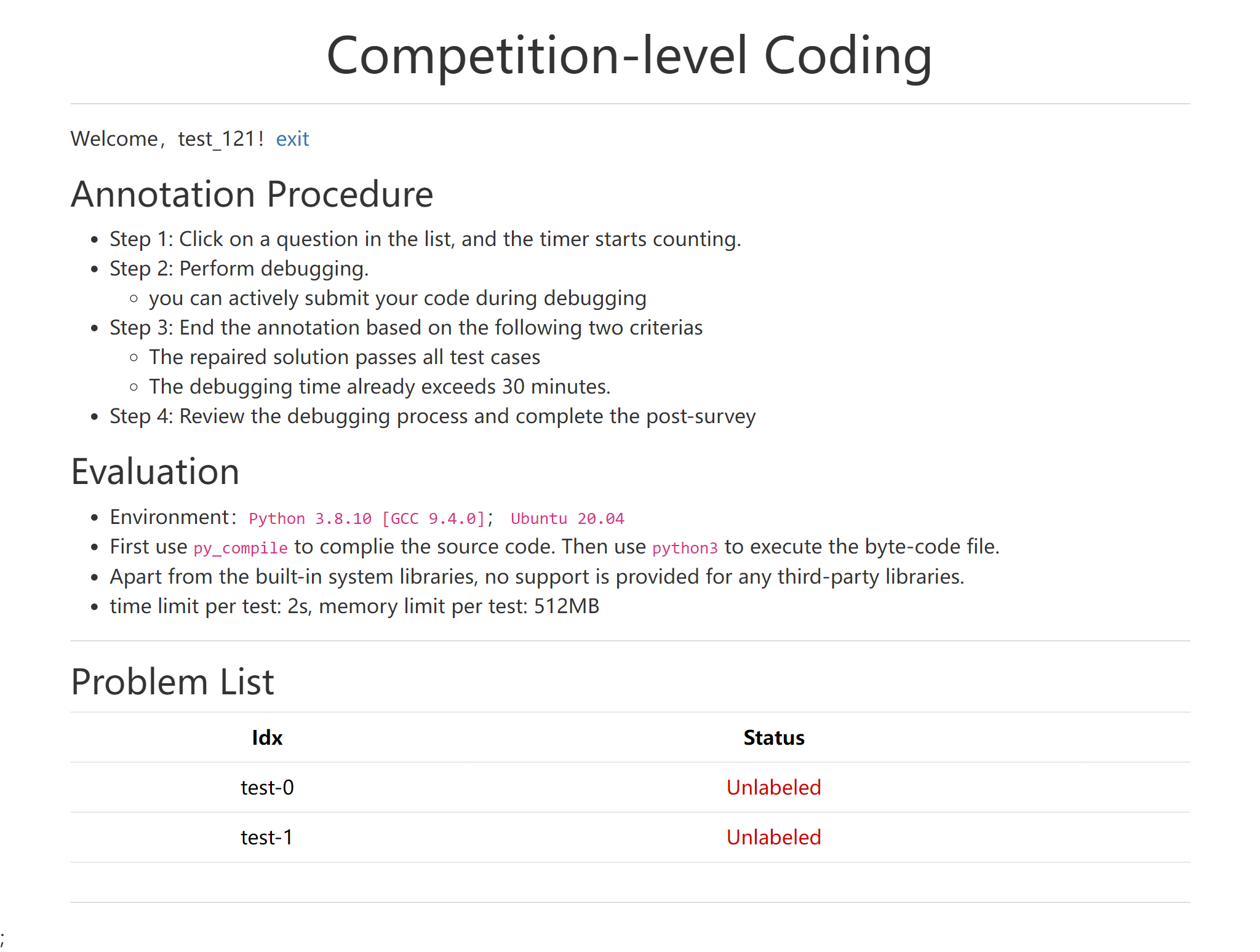}
         \caption{}
         \label{fig:interface-list}
     \end{subfigure}
\end{figure*}

\begin{figure*}[h]\ContinuedFloat
     \centering
     \begin{subfigure}[b]{1.0\textwidth}
         \centering
         \includegraphics[width=0.6\textwidth]{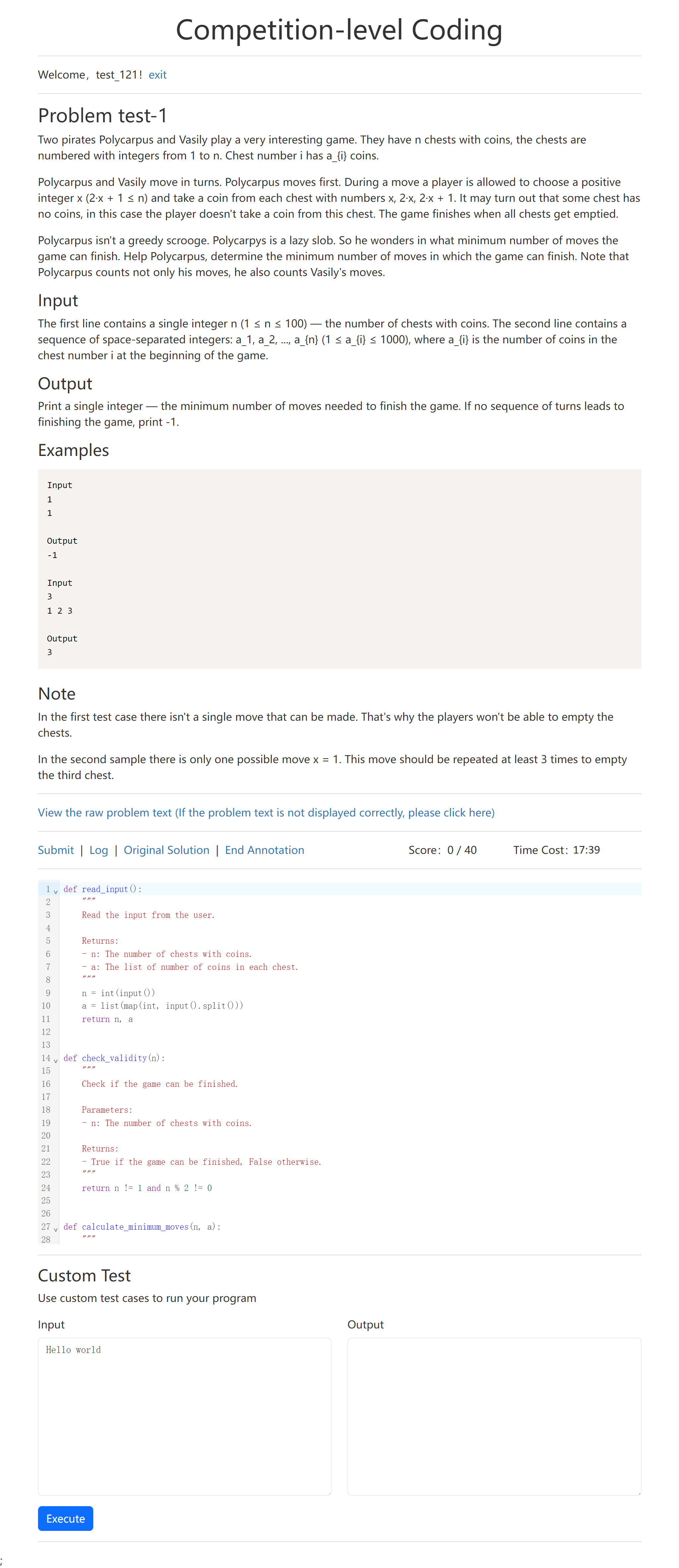}
         \caption{}
         \label{fig:interface-label}
     \end{subfigure}
        \caption{Screenshots of our labeling interface. (a) After entering the OJ system, labelers can see a brief labeling instruction and a problem list. (b) For each problem, annotators are required to perform the debugging task. During debugging, our OJ system supports annotators to run their custom test cases.}
        \label{fig:interface}
\end{figure*}

\section{Example Prompts} \label{appendix:prompt}

\subsection{Generating Initial Solution}

For generating the Initial solution, we adopt the prompt in Table \ref{tab:prompt_solution}.

\begin{table*}[h]
\centering
\small
\begin{tabular}{ |p{0.95\linewidth}| } 
\hline
\textbf{Problem}\\
You are given a tree consisting of $n$ vertices. A number is written on each vertex; the number on vertex $i$ is equal to $a_i$.

Let's denote the function $g(x, y)$ as the greatest common divisor of the numbers written on the vertices belonging to the simple path from vertex $x$ to vertex $y$ (including these two vertices). Also let's denote $dist(x, y)$ as the number of vertices on the simple path between vertices $x$ and $y$, including the endpoints. $dist(x, x) = 1$ for every vertex $x$.

Your task is to calculate the maximum value of $dist(x, y)$ among such pairs of vertices that $g(x, y) > 1$.

-----Input-----

The first line contains one integer $n$ — the number of vertices $(1 \le n \le 2 \cdot 10^5)$.

The second line contains $n$ integers $a_1$, $a_2$, ..., $a_n$ $(1 \le a_i \le 2 \cdot 10^5)$ — the numbers written on vertices.

Then $n - 1$ lines follow, each containing two integers $x$ and $y$ $(1 \le x, y \le n, x \ne y)$ denoting an edge connecting vertex $x$ with vertex $y$. It is guaranteed that these edges form a tree.

-----Output-----

If there is no pair of vertices $x, y$ such that $g(x, y) > 1$, print $0$. Otherwise print the maximum value of $dist(x, y)$ among such pairs...[TRUNCATED]\\
\textbf{Solution}
\begin{lstlisting}[language=Octave, basicstyle=\scriptsize]
import sys

answer = 1
z = True
primes = []
for i in range (2, 5 * 10 ** 2):
    v = True
    for p in primes:
        if i \% p == 0:
            v = False
    if v == True:
        primes.append(i)
n = int(sys.stdin.readline().strip())
a = list(map(int, sys.stdin.readline().strip().split()))
if sum(a) == n:
    z = False
for i in range (0, n):
    x = a[i]
    a[i] = []
    for p in primes:
        if x \% p == 0:
            a[i].append([p, 1]) \\
            x = x // p
            while x \% p == 0:
                x = x // p
    if x != 1:
        a[i].append([x, 1])
neighbours = [[] for i in range (0, n)]
for i in range (0, n - 1):
    line = sys.stdin.readline().strip().split()
    neighbours[int(line[0]) - 1].append(int(line[1]) - 1)
    neighbours[int(line[1]) - 1].append(int(line[0]) - 1)
leaves = []
for i in range (0, n):
    if len(neighbours[i]) == 1:
        leaves.append(i)
while len(leaves) > 1:
    x = leaves.pop()
    y = neighbours[x][0]
    neighbours[y].remove(x)
    if len(neighbours[y]) == 1:
        leaves.append(y)
    for p in a[x]:
        for q in a[y]:
            if p[0] == q[0]:
                answer = max([answer, p[1] + q[1]])
                q[1] = max([q[1],p[1]+1])
if z == False:
    print(0)
else:
    print(answer)
\end{lstlisting}
\textbf{Problem}\\
\{\{Problem\}\}\\
\hline
\end{tabular}
\caption{Excerpt of the input prompt for generating initial solutions.}
\label{tab:prompt_solution}
\end{table*}

\subsection{Vanilla Decomposition}

For vanilla decomposition, we adopt the prompt in Table \ref{tab:prompt_decomp}.

\begin{table*}[h]
\scriptsize
{
\begin{tabular}{|p{\linewidth}| } 
\hline
\textbf{Problem}\\
You are given a tree consisting of $n$ vertices. A number is written on each vertex; the number on vertex $i$ is equal to $a_i$.

Let's denote the function $g(x, y)$ as the greatest common divisor of the numbers written on the vertices belonging to the simple path from vertex $x$ to vertex $y$ (including these two vertices). Also let's denote $dist(x, y)$ as the number of vertices on the simple path between vertices $x$ and $y$, including the endpoints. $dist(x, x) = 1$ for every vertex $x$.

Your task is to calculate the maximum value of $dist(x, y)$ among such pairs of vertices that $g(x, y) > 1$...[TRUNCATED]\\
\textbf{Initial Solution}
\begin{lstlisting}[language=Octave, basicstyle=\tiny]
#...[TRUNCATED]
for i in range (2, 5 * 10 ** 2):
    v = True
    for p in primes:
        if i \% p == 0:
            v = False
    if v == True:
        primes.append(i)
n = int(sys.stdin.readline().strip())
a = list(map(int, sys.stdin.readline().strip().split()))
if sum(a) == n:
    z = False
for i in range (0, n):
    x = a[i]
    a[i] = []
    for p in primes:
        if x \% p == 0:
            a[i].append([p, 1])
            x = x // p
            while x \% p == 0:
                x = x // p
    if x != 1:
        a[i].append([x, 1])
#...[TRUNCATED]
\end{lstlisting}
Your task is to decompose the initial solution into separate functions that handle specific tasks. This will make the code more modular and easier to understand and debug.\\
\textbf{Decomposed Solution}
 \begin{lstlisting}[language=Octave, basicstyle=\tiny]
def sieve_of_eratosthenes(limit):
    """
    Generate a list of prime numbers up to the specified limit.
    Parameters:
    - limit: The upper bound up to which prime numbers are to be found.
    Returns:
    - primes: A list of prime numbers up to the specified limit.
    """
    primes = []
    is_prime = [True] * (limit + 1)
    for p in range(2, limit + 1):
        if is_prime[p]:
            primes.append(p)
            for i in range(p * p, limit + 1, p):
                is_prime[i] = False
    return primes

def factorize_numbers(a, primes):
    """
    Factorize each number in the list 'a' using the list of known primes.
    Parameters:
    - a: The list of numbers to be factorized.
    - primes: The list of known prime numbers.
    Returns:
    - factorized: A list of factorized numbers, where each number is represented as a list of prime factors and their exponents.
    """
    factorized = []
    for x in a:
        factors = []
        for p in primes:
            if x \% p == 0:
                while x \% p == 0:
                    x //= p
                factors.append((p, 1))
        if x != 1:
            factors.append((x, 1))
        factorized.append(factors)
    return factorized

def find_leaves(neighbours):
    """
    Find the leaf nodes (vertices with only one adjacent vertex) of a tree.
    Parameters:
    - neighbours: The adjacency list representing the tree.
    Returns:
    - leaves: A list of leaf nodes in the tree.
    """
    leaves = []
    for i, neighbours_list in enumerate(neighbours):
        if len(neighbours_list) == 1:
            leaves.append(i)
    return leaves
#...[TRUNCATED]
\end{lstlisting}
 \textbf{Problem}\\
\{\{Problem\}\}\\
\textbf{Initial Solution}\\
\{\{Initial Solution\}\}\\
 \hline
 \end{tabular}
 }
\caption{Excerpt of the input prompt for generating vanilla decomposition.}
\label{tab:prompt_decomp}
\end{table*}

\subsection{Critique}

For generating critique on decomposition, we adopt the prompt in Table \ref{tab:prompt_critic}.

\begin{table*}[h]
\scriptsize
{
\begin{tabular}{|p{\linewidth}| } 
\hline
\textbf{Problem}\\
Two pirates Polycarpus and Vasily play a very interesting game. They have n chests with coins, the chests are numbered with integers from 1 to n. Chest number i has $a_{i}$ coins. 

Polycarpus and Vasily move in turns. Polycarpus moves first. During a move a player is allowed to choose a positive integer $x (2·x + 1 \leq n)$ and take a coin from each chest with numbers $x, 2·x, 2·x + 1$. It may turn out that some chest has no coins, in this case the player doesn't take a coin from this chest. The game finishes when all chests get emptied...[TRUNCATED]
\\
\textbf{Initial Solution}
\begin{lstlisting}[language=Octave, basicstyle=\tiny]
def is_valid_input(n, a):
    """
    Check if the input is valid for the given constraints.

    Parameters:
    - n: The number of chests with coins.
    - a: The list of number of coins in each chest.

    Returns:
    - True if the input is valid, False otherwise.
    """
    return n % 2 == 1

def initialize_chests(a):
    """
    Initialize the list of chests with coins, adding extra elements at the beginning and end.

    Parameters:
    - a: The list of number of coins in each chest.

    Returns:
    - The modified list of chests.
    """
    return [0] + a + [0, 0]

def make_move(a, i):
    """
    Perform a move in the game, updating the number of coins in the chests accordingly.

    Parameters:
    - a: The list of number of coins in each chest.
    - i: The index of the chest to perform the move on.
    """
    m = min(a[i], a[i - 1])
    a[i // 2] = max(0, a[i // 2] - m)
    a[i] -= m
    a[i - 1] -= m

def calculate_minimum_moves(n, a):
    """
    Calculate the minimum number of moves needed to finish the game.

    Parameters:
    - n: The number of chests with coins.
    - a: The list of number of coins in each chest.

    Returns:
    - The minimum number of moves needed to finish the game.
    """
    if n == 1 or not is_valid_input(n, a):
        return -1
    a = initialize_chests(a)
    for i in range(n, 0, -2):
        make_move(a, i)
    return sum(a)
#...[TRUNCATED]
\end{lstlisting}

\textbf{Critique Request}\\
Identify specific ways in which the decomposition strategy of the initial solution is less effective in assisting humans to debug the solution. \\

\textbf{Critique}\\
1. Decomposition should be around logical breakpoints, but not syntax. However, \inlinecodesmall{make\_move} and \inlinecodesmall{initialize\_chests} are not the main logic steps of this problem. We should present a clear high-level logic that aligns with how humans reason about programs conceptually. In this problem, the main logic steps are input reading, checking game finishing conditions, calculating minimum steps, and output printing.\\
2. Boundary conditions (i.e., \inlinecodesmall{if n == 1 and n \% 2 == 1} should be explicitly highlighted as individual functions in the main logic, instead of in the function \inlinecodesmall{calculate\_minimum\_moves}.\\
(...TRUNCATED)\\
 \textbf{Problem}\\
\{\{Problem\}\}\\
\textbf{Initial Solution}\\
\{\{Initial Solution\}\}\\
 \hline
 \end{tabular}
 }
\caption{Excerpt of the input prompt for generating critique on decomposition}
\label{tab:prompt_critic}
\end{table*}

\subsection{Refine}

For generating the refined decomposed solution, we adopt the prompt in Table \ref{tab:prompt_refine}.

\begin{table*}[h]
\scriptsize
{
\begin{tabular}{|p{\linewidth}| } 
\hline
\textbf{Problem}\\
Two pirates Polycarpus and Vasily play a very interesting game. They have n chests with coins, the chests are numbered with integers from 1 to n. Chest number i has $a_{i}$ coins. 

Polycarpus and Vasily move in turns. Polycarpus moves first. During a move a player is allowed to choose a positive integer $x (2·x + 1 \leq n)$ and take a coin from each chest with numbers $x, 2·x, 2·x + 1$. It may turn out that some chest has no coins, in this case the player doesn't take a coin from this chest. The game finishes when all chests get emptied...[TRUNCATED]
\\
\textbf{Initial Solution}
\begin{lstlisting}[language=Octave, basicstyle=\tiny]
def is_valid_input(n, a):
    """Check if the input is valid for the given constraints...(TRUNCATED)"""
    return n % 2 == 1

def initialize_chests(a):
    """Initialize the list of chests with coins, adding extra elements at the beginning and end...(TRUNCATED)"""
    return [0] + a + [0, 0]

def make_move(a, i):
    """Perform a move in the game, updating the number of coins in the chests accordingly...(TRUNCATED)"""
    m = min(a[i], a[i - 1])
    a[i // 2] = max(0, a[i // 2] - m)
    a[i] -= m
    a[i - 1] -= m

def calculate_minimum_moves(n, a):
    """Calculate the minimum number of moves needed to finish the game...(TRUNCATED)"""
    if n == 1 or not is_valid_input(n, a):
        return -1
    a = initialize_chests(a)
    for i in range(n, 0, -2):
        make_move(a, i)
    return sum(a)

def main():
    n = int(input())
    a = list(map(int, input().split()))
    result = calculate_minimum_moves(n, a)
    print(result)
#...[TRUNCATED]
\end{lstlisting}

\textbf{Critique}\\
1. Decomposition should be around logical breakpoints, but not syntax. However, \inlinecodesmall{make\_move} and \inlinecodesmall{initialize\_chests} are not the main logic steps of this problem. We should present a clear high-level logic that aligns with how humans reason about programs conceptually. In this problem, the main logic steps are input reading, checking game finishing conditions, calculating minimum steps, and output printing.\\
2. Boundary conditions (i.e., \inlinecodesmall{if n == 1 and n \% 2 == 1} should be explicitly highlighted as individual functions in the main logic, instead of in the function \inlinecodesmall{calculate\_minimum\_moves}.\\\\

Refine the initial solution based on the above critique. Do not introduce modification unrelated to the critique.\\
\textbf{Refine}
\begin{lstlisting}[language=Octave, basicstyle=\tiny]
...(TRUNCATED)
def check_validity(n):
    """Check if the game can be finished...(TRUNCATED)"""
    return n != 1 and n % 2 != 0

def calculate_minimum_moves(n, a):
    """Calculate the minimum number of moves needed to finish the game...(TRUNCATED)"""
    a = [0] + a + [0, 0]
    for i in range(n, 0, -2):
        m = min(a[i], a[i - 1])
        a[i // 2] = max(0, a[i // 2] - m)
        a[i] -= m
        a[i - 1] -= m
    return sum(a)

def main():
    n, a = read_input()
    if not check_validity(n):
        print(-1)
    else:
        print(calculate_minimum_moves(n, a))
#...[TRUNCATED]
\end{lstlisting}
(...TRUNCATED)\\
 \textbf{Problem}\\
\{\{Problem\}\}\\
\textbf{Initial Solution}\\
\{\{Initial Solution\}\}\\
\textbf{Critique}\\
\{\{Critique\}\}\\
 \hline
 \end{tabular}
 }
\caption{Excerpt of the input prompt for generating refined decomposition}
\label{tab:prompt_refine}
\end{table*}

\subsection{Rank}
For ranking which decomposition leads to higher human efficiency, we adopt the prompt in Table \ref{tab:prompt_rank}.

\begin{table*}[h]
\scriptsize
{
\begin{tabular}{|p{\linewidth}| } 
\hline
\textbf{Problem}\\
Two pirates Polycarpus and Vasily play a very interesting game. They have n chests with coins, the chests are numbered with integers from 1 to n. Chest number i has $a_{i}$ coins. 

Polycarpus and Vasily move in turns. Polycarpus moves first. During a move a player is allowed to choose a positive integer $x (2·x + 1 \leq n)$ and take a coin from each chest with numbers $x, 2·x, 2·x + 1$. It may turn out that some chest has no coins, in this case the player doesn't take a coin from this chest. The game finishes when all chests get emptied...[TRUNCATED]
\\
\textbf{Decomposed Solution A}
\begin{lstlisting}[language=Octave, basicstyle=\tiny]
def is_valid_input(n, a):
    """Check if the input is valid for the given constraints...(TRUNCATED)"""
    return n % 2 == 1

def initialize_chests(a):
    """Initialize the list of chests with coins, adding extra elements at the beginning and end...(TRUNCATED)"""
    return [0] + a + [0, 0]

def make_move(a, i):
    """Perform a move in the game, updating the number of coins in the chests accordingly...(TRUNCATED)"""
    m = min(a[i], a[i - 1])
    a[i // 2] = max(0, a[i // 2] - m)
    a[i] -= m
    a[i - 1] -= m

def calculate_minimum_moves(n, a):
    """Calculate the minimum number of moves needed to finish the game...(TRUNCATED)"""
    if n == 1 or not is_valid_input(n, a):
        return -1
    a = initialize_chests(a)
    for i in range(n, 0, -2):
        make_move(a, i)
    return sum(a)

def main():
    n = int(input())
    a = list(map(int, input().split()))
    result = calculate_minimum_moves(n, a)
    print(result)
#...[TRUNCATED]
\end{lstlisting}

\textbf{Decomposed Solution B}
\begin{lstlisting}[language=Octave, basicstyle=\tiny]
...(TRUNCATED)
def check_validity(n):
    """Check if the game can be finished...(TRUNCATED)"""
    return n != 1 and n % 2 != 0

def calculate_minimum_moves(n, a):
    """Calculate the minimum number of moves needed to finish the game...(TRUNCATED)"""
    a = [0] + a + [0, 0]
    for i in range(n, 0, -2):
        m = min(a[i], a[i - 1])
        a[i // 2] = max(0, a[i // 2] - m)
        a[i] -= m
        a[i - 1] -= m
    return sum(a)

def main():
    n, a = read_input()
    if not check_validity(n):
        print(-1)
    else:
        print(calculate_minimum_moves(n, a))
#...[TRUNCATED]
\end{lstlisting}

\textbf{Evaluation Request}
Given two decomposed solutions A and B, determine which one of them is more effective in assisting humans to debug the solution.

\textbf{Evaluation Results}
Decomposed Solution B is better.\\
1. Decomposition should be around logical breakpoints, but not syntax. However, \inlinecodesmall{make\_move} and \inlinecodesmall{initialize\_chests} in Decomposed Solution A are not the main logic steps of this problem. We should present a clear high-level logic that aligns with how humans reason about programs conceptually. In this problem, the main logic steps are input reading, checking game finishing conditions, calculating minimum steps, and output printing, as shown in Decomposed Solution B.\\
2. Boundary conditions (i.e., \inlinecodesmall{if n == 1 and n \% 2 == 1}) should be explicitly highlighted as individual functions in the main logic, while Decomposed Solution A places it in the \inlinecodesmall{calculate\_minimum\_moves} function.\\
(...TRUNCATED)\\

 \textbf{Problem}\\
\{\{Problem\}\}\\
\textbf{Decomposed Solution A}\\
\{\{Decomposed Solution A\}\}\\
\textbf{Decomposed Solution B}\\
\{\{Decomposed Solution B\}\}\\
 \hline
 \end{tabular}
 }
\caption{Excerpt of the input prompt for ranking different decompositions.}
\label{tab:prompt_rank}
\end{table*}

\section{Case Study}

\begin{figure*}[t!]
    \centering
    \includegraphics[width=0.8\linewidth]{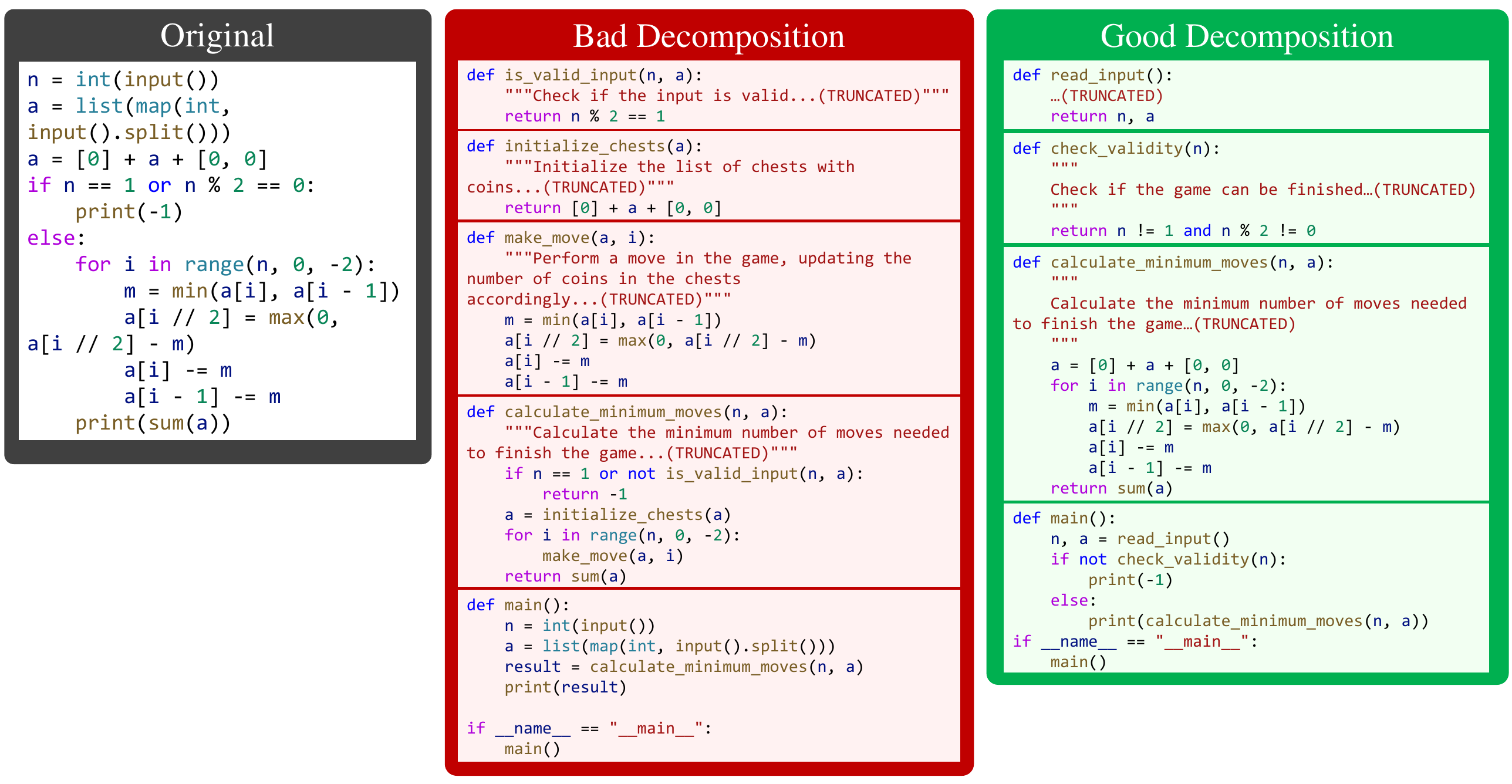}
    \caption{Excerpt of the decomposition that highlights boundary conditions in the training data.}
    \label{fig:case_boundary_condition_1}
\end{figure*}

\begin{figure*}[t!]
    \centering
    \includegraphics[width=0.9\linewidth]{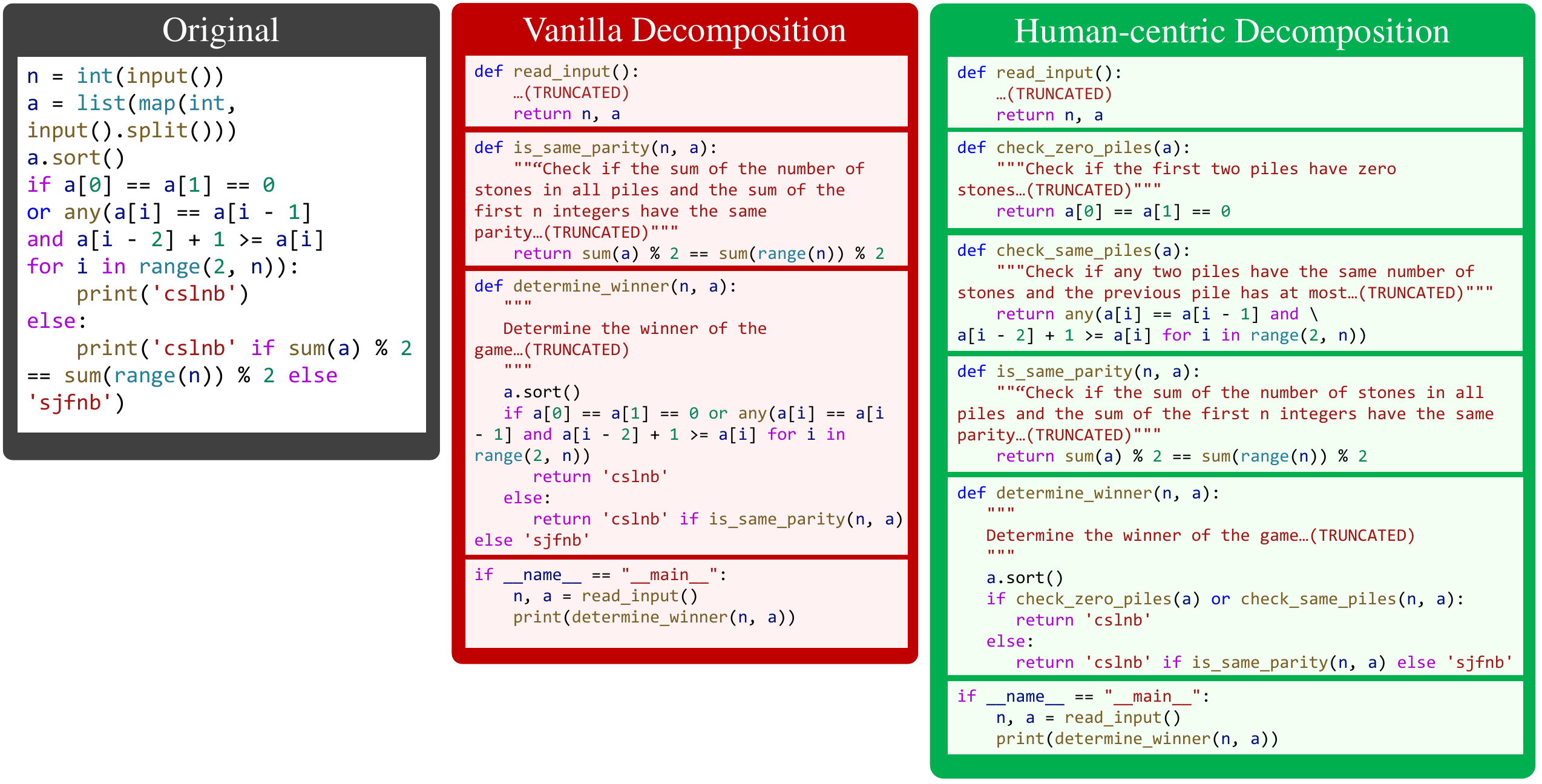}
    \caption{Excerpt of the decomposition that highlights boundary conditions in the test data.}
    \label{fig:case_boundary_condition_2}
\end{figure*}

\begin{figure*}[t!]
    \centering
    \includegraphics[width=\linewidth]{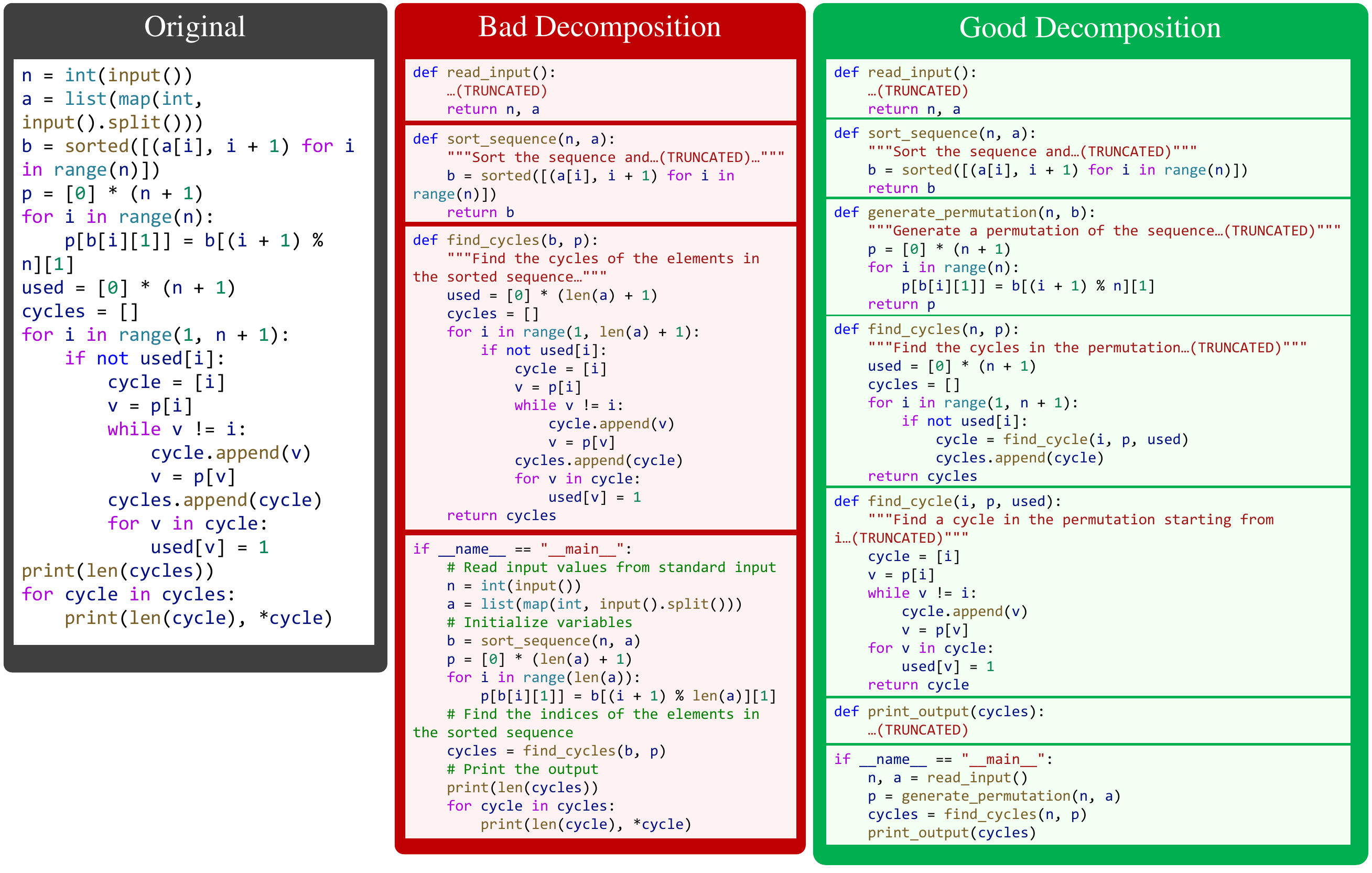}
    \caption{Excerpt of the decomposition that creates simpler subtasks in the training data.}
    \label{fig:case_simple_subtask_1}
\end{figure*}

\begin{figure*}[t!]
    \centering
    \includegraphics[width=0.9\linewidth]{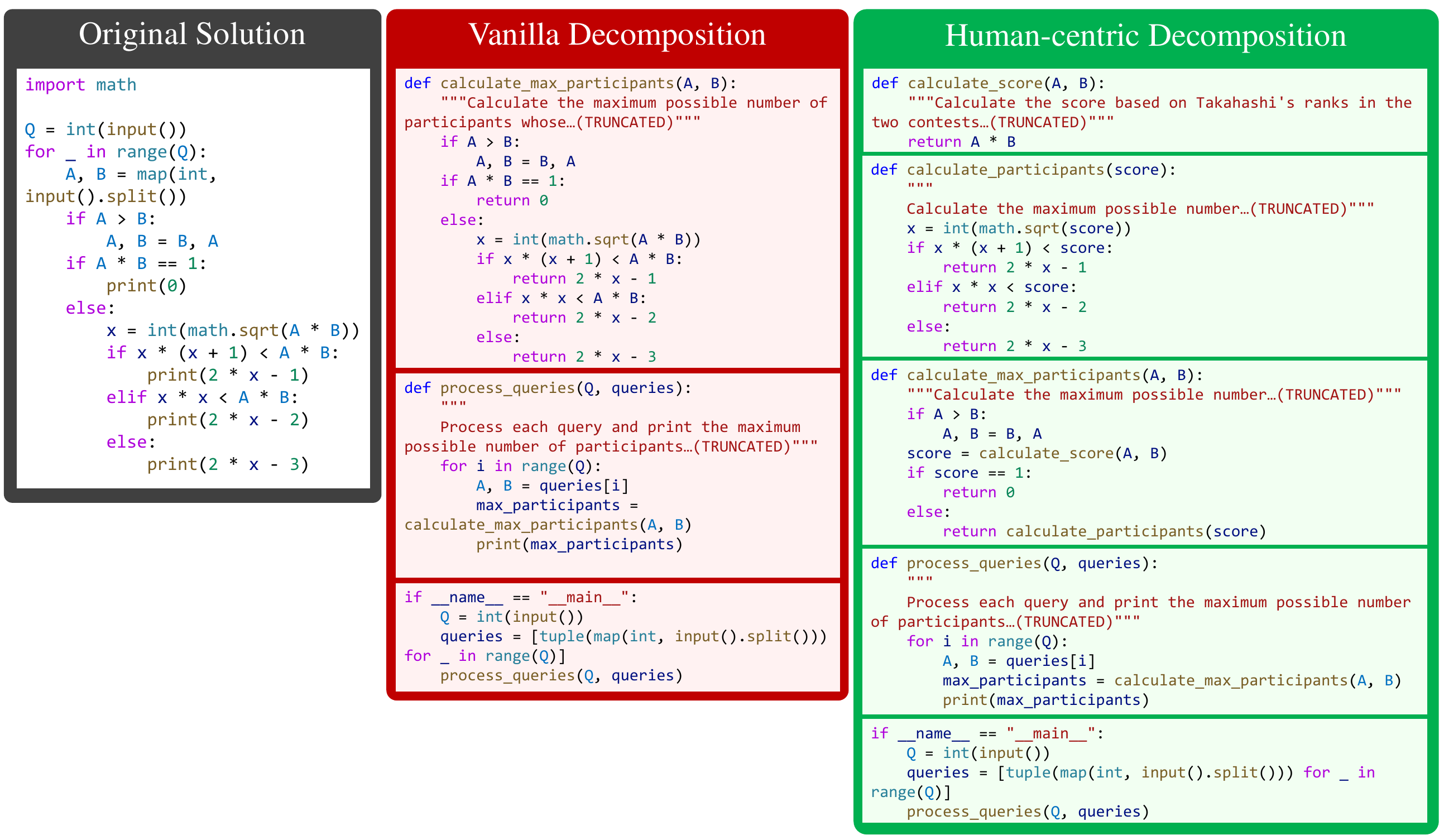}
    \caption{Excerpt of the decomposition that creates simpler subtasks in the test data}
    \label{fig:case_simple_subtask_2}
\end{figure*}

\begin{figure*}[t!]
    \centering
    \includegraphics[width=\linewidth]{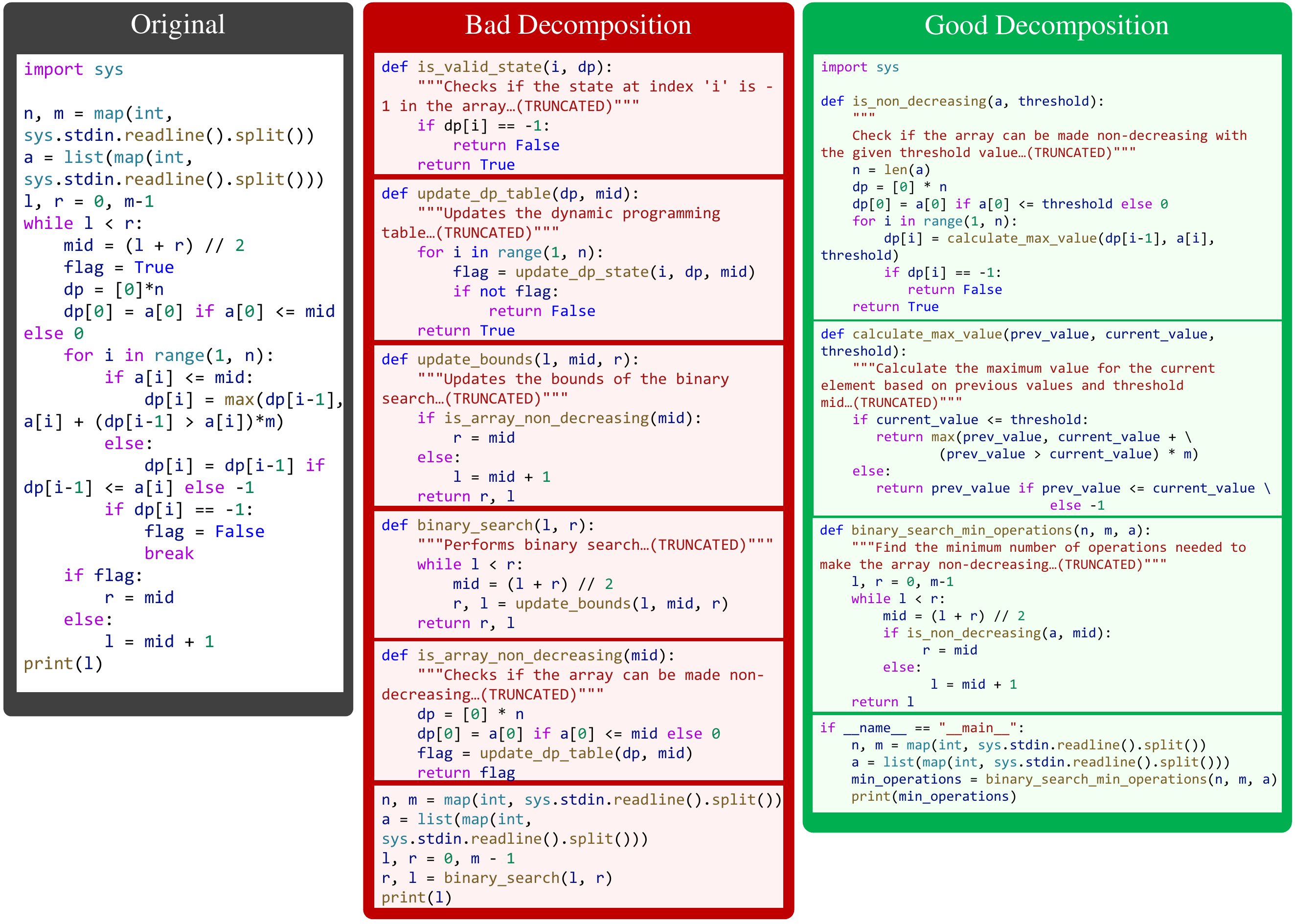}
    \caption{Excerpt of the decomposition that presents clear high-level logic in the training data.}
    \label{fig:case_high_level_1}
\end{figure*}

\begin{figure*}[t!]
    \centering
    \includegraphics[width=\linewidth]{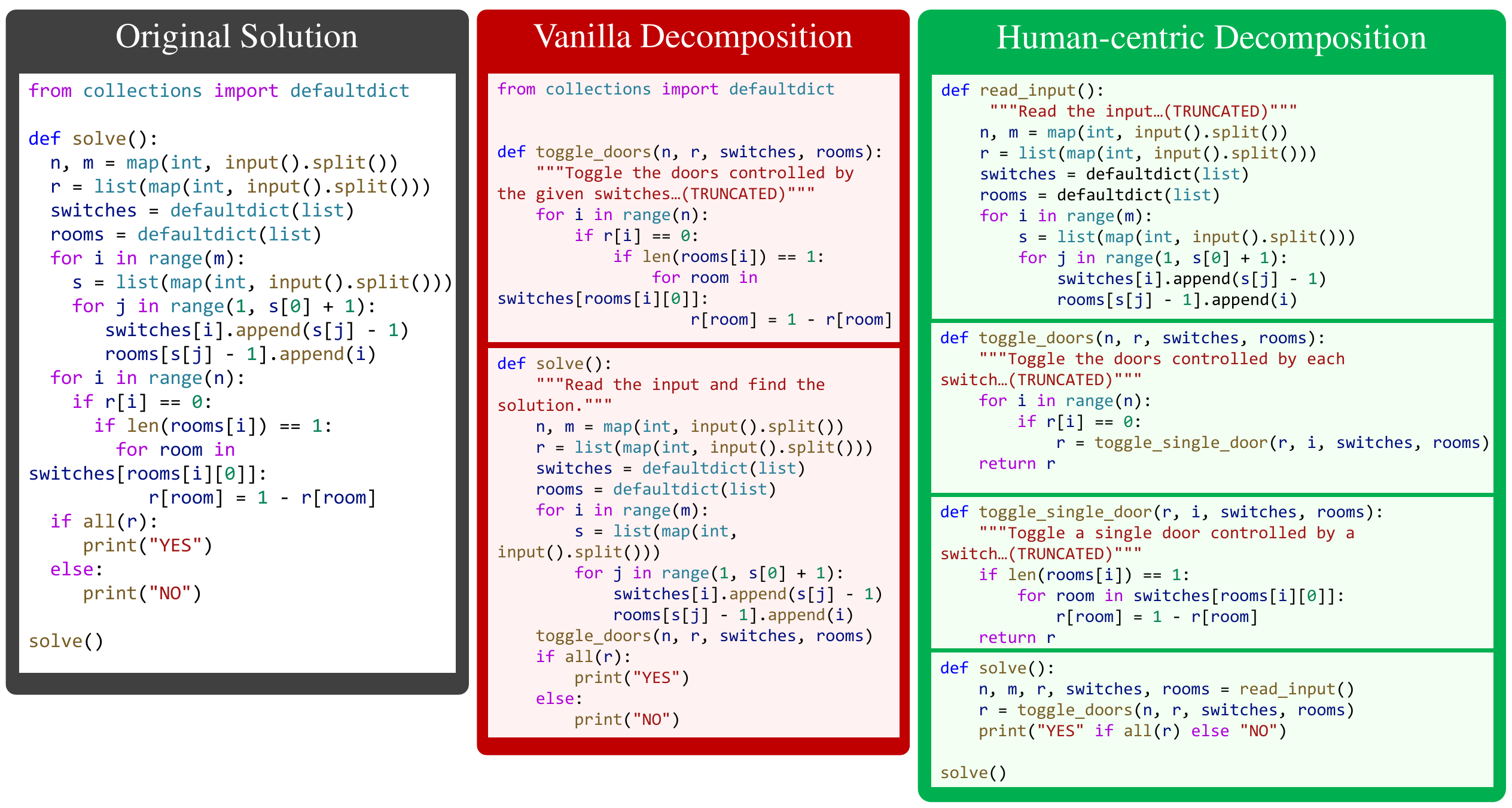}
    \caption{Excerpt of the decomposition that presents clear high-level logic in the test data.}
    \label{fig:case_high_level_2}
\end{figure*}

We present several cases to illustrate how decomposition assists humans in practice by highlighting boundary conditions (Figure \ref{fig:case_boundary_condition_1}, Figure \ref{fig:case_boundary_condition_2}), creating simpler and more managable subtasks (Figure \ref{fig:case_simple_subtask_1}, Figure \ref{fig:case_simple_subtask_2}), and presenting clear high-level logic (Figure \ref{fig:case_high_level_1}, Figure \ref{fig:case_high_level_2}).

\end{document}